\documentclass[twoside,11pt]{article}
\usepackage{amsmath}
\usepackage{hyperref}
\usepackage{unicode}
\usepackage{eurosym}
\usepackage{graphicx}
\usepackage{jair, theapa, rawfonts}
\usepackage{algorithm}
\usepackage{algpseudocode}
\usepackage{subcaption}
\usepackage{pgfplots}
\pgfplotsset{compat=1.13}
\usetikzlibrary{pgfplots.groupplots}
\usepackage{pgfplotstable}

\usepackage{booktabs}

\ShortHeadings{Satisficing Mentalizing}
{P\"oppel \& Kopp}

\newcommand{\meanAndStdColumn}[1]{
    \pgfplotstableset{
        create on use/#1 and Std/.style={
            create col/assign/.code={
                \edef\entry{\noexpand\pgfmathprintnumber[fixed zerofill, precision=4]{\thisrow{#1}} (\noexpand\pgfmathprintnumber[fixed zerofill, precision=2]{\thisrow{#1std}})}
                \pgfkeyslet{/pgfplots/table/create col/next content}{\entry}
            }
        }
    }
}

\newcommand{\accuracyTable}[3]{
    \pgfplotstableread[col sep=tab,]{#1}\tableData
    \begin{table}[h]
    	\centering
        \meanAndStdColumn{overall}
        \meanAndStdColumn{twg}
        \meanAndStdColumn{tw}
        \meanAndStdColumn{tg}
    	\pgfplotstabletypeset[
    	columns={Model, overall and Std, twg and Std, tw and Std, tg and Std},
     	every head row/.style={
     		before row={
     			\toprule
    		},
    		after row=\midrule,
    	},
    	every last row/.style={
    		after row=\bottomrule},
    	every first column/.style={
    		column type/.add={}{|}
    	},
     	columns/overall and Std/.style       ={column name=Overall},
        columns/twg and Std/.style       ={column name=NU},
        columns/tw and Std/.style       ={column name=DU},
        columns/tg and Std/.style       ={column name=PU},
    	col sep=&,row sep=\\,string type,]{\tableData}
    	\caption{#3}
    	\label{#2}
    \end{table}
}

\begin{document}

\title{Satisficing Mentalizing: Bayesian Models of Theory of Mind Reasoning in Scenarios with Different Uncertainties}

\author{\name Jan P\"oppel \email jpoeppel@techfak.uni-bielefeld.de \\
       \name Stefan Kopp \email skopp@techfak.uni-bielefeld.de \\
       \addr Bielefeld University, Inspiration 1\\
       33619 Bielefeld, Germany}


\maketitle

\begin{abstract}
The ability to interpret the mental state of another agent based on its behavior, also called Theory of Mind (ToM), is crucial for humans in any kind of social interaction. 
Artificial systems, such as intelligent assistants, would also greatly benefit from such mentalizing capabilities. 
However, humans and systems alike are bound by limitations in their available computational resources. This raises the need for \emph{satisficing} mentalizing, reconciling accuracy and efficiency in mental state inference that is \emph{good enough} for a given situation. In this paper, we present  different Bayesian models of ToM reasoning and evaluate them based on actual human behavior data that were generated under different kinds of uncertainties. We propose a \emph{Switching} approach that combines specialized models, embodying simplifying presumptions, in order to achieve a more statisficing mentalizing compared to a \emph{Full Bayesian ToM} model.
\footnote{This paper is an extended version of \cite{poppel2018satisficing}.}
\end{abstract}

\section{Introduction}
Artificial systems are becoming increasingly widespread in our everyday lives and interact with us in various contexts. They provide information or media entertainment via spoken language interaction in home environments, assist in driving or navigating in traffic, or collaborate during manufacturing tasks. In all of these scenarios, one prerequisite for artificial systems to effectively provide support to their users is a capability of ``understanding'' human action and interaction, at least to a certain extent. 

Humans have a remarkable ability to infer hidden information about another 
one's actions such as desires, potentially false beliefs, preferences, and even emotions, only from observing their behavior \cite{wellman2004scaling}. This ability is famously called "Theory of Mind" (ToM) \cite{premack1978does}. For example, if we see somebody walking towards the office kitchen with a mug in her hand, we can quite confidently infer that she is looking for a new coffee and that she believes that there is coffee to be found in the kitchen. This will also hold if we know that the coffee just run out, because we are able to attribute beliefs that differ from our own knowledge. This "mentalizing" ability is assumed to develop gradually as we grow up, with the inference of another one's desires coming first before we are able to infer more complex mental states such as (false) beliefs, preferences, and even emotions \cite{wellman2004scaling}. Generally, children are assumed to have developed their all of their ToM abilities, albeit not necessarily to the fullest extend, by the age of 6 \cite{flavell1999cognitive,moran2013lifespan}.
At that point, children can not only infer another agent's potentially false belief, but also employ recursive reasoning about what the other thinks about oneself or what oneself might think about them \cite{grueneisen2015know}.
Our ToM capabilities help us in all our social interactions, from learning more about our environment when observing others interacting with it, over allowing us better cooperation, by taking our partner's perspective into account, to being able to trick or deceive others gaining us an advantage in competitive settings. 


While humans often use their mentalizing abilities subconsciously, they do not always employ them to their fullest extend \cite{keysar2003limits}. 
\citeA{de2015higher} have shown that we usually only employ first or second order ToM reasoning in certain situations, even though we are capable of up to the sixth or seventh order.

A prominent example for this are egocentric tendencies found in communication \cite{keysar2007communication}. \citeA{keysar2007communication} showed that people often do not correctly take their communication partner's perspective into account, but rather project their own egocentric views onto them.
They contribute this lack of perspective taking at least in part to working memory limitations, highlighting that mentalizing of more complex mental states is demanding even for adults. 

In fact, it has long been argued that humans are not capable of objective rationality, but can only follow a \emph{bounded rationality} \cite{simon1955behavioral,aumann1997rationality} due to the limitations of our own (physical and mental) capabilities. 
Due to these limitations we are usually employing different kinds of heuristics to solve inference and decision making tasks.
\citeA{simon1955behavioral} argued that humans strife to be \emph{satisficing}, meaning that we usually choose options that are \emph{good enough} to satisfy a given need instead of actually evaluating all possible options in order to choose the objectively best one.

The most prominent approach in recent years to model ToM for artificial systems has been the Bayesian Theory of Mind (BToM) framework by \citeA{Baker2009,baker2017rational} which formalizes an inverse planning approach into a Bayesian framework. 
Within the BToM framework an agent's behavior is modeled by a generative probabilistic model specifying how observable actions are caused by hidden mental states such as desires and/or beliefs.
Using Bayes' rule, one can invert these models to infer likely mental states based on observations. 
While this approach has been shown to make inferences correlating well with human judgments in a range of different scenarios \cite{UllmanTomer2009,jern2017people,velez2017interpreting}, its key problem is that Bayesian reasoning becomes prohibitively expensive for complex models, resulting in fine-tuned systems specialized to infer mental states in very specific situations. 

We argue for the need to consider \emph{satisficing} mentalizing for artificial systems in this paper. 
Contrary to most previous work on BToM models, we evaluate different models following the Bayesian Theory of Mind framework on actual human behavior data generated under different mental conditions with respect to their predictive accuracy as well as their computational efficiency.
We chose to evaluate the model's predictive accuracy of human behavior in a ToM setting primarily for two reasons: 
Firstly, previous research has already shown that the BToM framework is able to infer mental states that correlate well with inferences made by humans when considering handcrafted stimuli. For this work, we therefore assume that the choice of BToM models is appropriate for inferring mental states from behavior. However it has not been evaluated on actual, potentially noise, human behavior data.
Secondly, we believe that, at least for the domain chosen in this work, one cannot make good predictions regarding an agent's future actions without a sufficiently accurate account of said agent's mental state, as different mental state assumptions will lead to very different action predictions.
For these reasons, we believe the predictive accuracy to be a good metric for the quality of the mental state inference for our models.
We further propose a Switching strategy as a possible way towards \emph{satisficing} mentalizing within the Bayesian Theory of Mind framework. 
By only making use of more complex models when simpler models no longer suffice, we explore how an artificial system can make use of these different heuristics in order to perform mental reasoning \emph{good enough}, similar to humans.

To collect the behavioral data we performed an online study in which participants had to navigate a 2D gridworld while manipulating the information available to them, thus inducing different uncertainties.
We argue that these different conditions influenced the decision making process of our participants in distinct ways, resulting in the need for different specialized BToM models to best explain the observable behavior.
We will show that an alternative \emph{Full BToM} model, which tries to model all possible mental states concurrently, quickly becomes prohibitively expensive to evaluate (exactly) while not necessarily performing 
accurately.
Instead we propose to \emph{switch} between different specialized models (or heuristics if you will), which not only reduces the computational costs, but also shows superior predictive accuracy, potentially bringing it closer to the \emph{satisficing} mentalizing resulting from the bounded rationality in humans.

The rest of this paper is structured as follows: First we will give a brief summary of the relevant literature regarding mentalizing for action understanding using the BToM framework as well as relevant findings in Cognitive Science. 
Afterwards we will present the aforementioned navigation scenario as well as the different conditions we employed there in order to induce behavior under different kinds of uncertainty.
We will present the data we collected and discuss the different behavior patterns we observed.
In section \ref{sec:models} we introduce the different computational models for action understanding 
formally before discussing results of their application and comparison on the original study data.

\section{Related Work}

There has been a lot of research on both bounded rationality as well as different approaches to action understanding and intention or plan recognition across a range of different domains. We will review some of the relevant findings in the following.

\subsection{Classical Approaches to Behavior Understanding}

Action, intention and plan recognition has been studied extensively in autonomous robots that need to be able to interpret the behavior of agents around them in order to function better but also in communication where an interlocutors intention was of interest. 
From there other domains such as entertainment and education have also come into the focus of researchers.

\citeA{bosse2007two} proposed a two-level \emph{Belief, Desire, Intention} (BDI) \cite{Rao1995} model for artificial agents. In this way, the classical BDI framework is used to create an agent that reasons about the mentalizing process of another agent, therefore performs Theory of Mind. 
This approach shares the classical benefits and limitations of the BDI framework: 
It provides well defined models that can be tailored to specific situations, which however is simultaneously its biggest weakness since the rules it follows need to be handcrafted and are usually not flexible enough to deal with variations in the scenario.

An early probabilistic approach to plan recognition has been proposed in \cite{charniak1993bayesian}. The authors stress the importance of a probabilistic consideration compared to the classical logic-based approaches and present rules for translating a plan recognition problem into a Bayesian network. 
\citeA{albrecht1998bayesian} also used Dynamic Bayesian Networks in order to perform plan recognition in an adventure game.
Multiple Hidden Markov Models (HMMs) are used in \citeA{han2000automated} in order to model robot behavior. Similar to our proposed switching approach, their recognition algorithm makes use of specialized models (HMMs in their case), each designed for a specific behavior. However, similar to word recognition and unlike our approach, all HMMs need to be evaluated for every new observation in order to figure out which HMMs accept the observations.

\citeA{pynadath2005psychsim} developed ``PsychSim, an implemented multiagent-based simulation tool for modeling interactions and influence'' (p. 1) 
to simulate school violence and bullying. PsychSim directly models beliefs about other agents using a POMDP where each agent does not only hold beliefs about the world, but also about the beliefs and assumed preferences of other agents. 
However, in order to save computational resources, \citeA{pynadath2005psychsim} do not model deeper recursive believes as in agent A believes that agent B beliefs that agent A knows X.
Conceptually, PsychSim has many similarities to the Bayesian Theory of Mind framework, as it formalizes the belief dynamics in a probabilistic framework (the POMDP). 
However in its presented state, it only allows for forward simulation of agents that take other agents into account.
It has not been used to infer the mental states of other agents based on their behavior.

An interesting intention-focused approach uses force dynamics to model the interaction between multiple agents that take each other into account \cite{crick2010controlling}. Their system is able to infer relationships between the different agents, such as that one chases the other, as well as to learn how its own actions influence the environment.
While their approach does not directly model more complex mental states such as beliefs, it can be seen as an example that simpler systems, relying on rudimentary sensory information, can achieve very sophisticated behavior.

Recently, \citeA{rabinowitz2018machine} present a ToM model based on deep learning. They combine deep neural networks trained to embed observable behavior of an agent navigating a maze into a low dimensional space which should represent the mental state of that agent.
They are then able to predict the agent's next actions from this mental state as well as an agent-specific character embedding using a prediction network.
While they cannot describe the observed agent's mental state clearly, they consistently find structure within the embedding space which can be related to certain mental states. 
However, at this point this system still requires a large amount of synthetic training data from artificial agents, which makes it currently unusable for the use with human behavior.

\subsection{Bayesian Theory of Mind}

The Bayesian Theory of Mind (BToM) framework proposed by \citeA{Baker2011} can be seen as a special case of inverse planning \cite{baker2007goal,Baker2009}. The premise that allows one to infer correct mental states from observations despite the infinite number of possible explanations is the \emph{principle of rationality}, i.e. the assumption that rational agents behave optimally \cite{dennett1989intentional}.
Under this assumption, one can assume that an agent's behavior maximizes its utility.
By inverting the planning problem we can thus infer an agent's goals or intentions in the form of its utility function.
The BToM framework directly encodes beliefs or other mental states as probability distributions \cite{Baker2011}. 
Their work has been greatly inspired by the nested belief representation and inference for multi-agent environments in \cite{zettlemoyer2009multi}.

Informally, the BToM framework can be described as follows:
By defining a generative model which relates mental states such as intentions and beliefs to actual actions, e.g. as a (partially observable) Markov Decision Problem (POMDP), an agent's actions can be predicted, given these mental states. 
Bayes' rule allows us to invert this model in order to infer these mental states from observations:

\begin{equation}
	\label{eq:btom}
	P(M|{Actions},{Environment}) \propto P({Actions}|M,{Environment})P(M|{Environment})
\end{equation}

with $M$ standing for our considered mental states (slightly adapted from \cite{Baker2009}).
This general approach, not always under the name of BToM, has been used successfully in a range of different scenarios, such as navigation \cite{baker2017rational}, compositional desires \cite{velez2017interpreting}, social interactions \cite{UllmanTomer2009} or inferring people's preferences \cite{jern2017people}.
The success of these models has usually been determined by running studies that collect human ratings on the mental states of interest in a range of generated stimuli and computing the correlation between these human ratings and the models' predictions. 
While these works have shown that the proposed models can make inferences similar to humans in those situations, these models were designed specifically for those scenarios with their parameters often fit to maximize correlation. 
Past research has not explored how well these models perform with a given set of parameters when the situation they are considering changes or certain assumptions are violated.

\subsection{Rational Reasoning Under Constrained Resources} 

The notion of rationality has long been studied from many different perspectives. 

\citeA{simon1955behavioral} was among the first to systematically cover the discrepancy between objective rationality and rationality with respect to the limitations of a living organism. 

These constraints include purely physical ones, such as the speed at which we can move, but also cognitive limitations, such as the size of our working memory.
\citeA{simon1955behavioral} argues that all these constraints, external as well as internal, need to be kept in mind when considering rationality and that humans can only ever achieve \emph{bounded rationality}.

Since then, there has been an increasing amount of work regarding this notion of bounded rationality, especially in the Cognitive Science community, with a special focus on discovering these limitations in greater detail. 
For those interested, a review regarding multiple cases of bounded rationality in humans can be found in \cite{jones1999bounded}. 
Furthermore, \citeA{haselton2015evolution} argue how many biases now attributed to bounded rationality may have evolved in order to improve our overall fitness. 

A very recent framework trying to formalize bounded rationality for the general case has been proposed by \citeA{lieder_griffiths_2019} with their ``Resource-rational analysis''. They formalize how humans may still maximize their expected utility, i.e. behave rationally, while being restricted by limited (computational) resources. A result of their framework is actually the emergence and solidification of heuristics, as applying (learned) heuristics will usually yield good results with comparatively little computational costs.
This is in line with the switching approach we are proposing in this paper, as it also makes use of given (e.g. previously learned) heuristics in the form of simplified models but still selects which one to choose. 

When considering the BToM framework, one also quickly recognises the need for more satisficing solutions that take resource constraints into account. 
As \citeA{Baker2009} already stated in their early work on the BToM framework:
``..,the inverse problem is ill-posed. Its solution requires strong prior knowledge of the structure and content of agents' mental states, and the ability to search over and evaluate a potentially very large space of possible mental state interpretations.'' (p. 330) 

While the strong prior knowledge requirement might potentially be mitigated somewhat using hierarchical learning approaches (e.g. \cite{Diaconescu2014}) in the future, the ``ability to search over and evaluate a potentially very large space..''
is a mayor challenge for Bayesian methods in general. 
Especially the normalization required in Bayes' rule forces one to integrate over all possible mental states, which quickly becomes intractable. 
Previously, this usually resulted in the consideration of only a small mental state space, coarse discretization of continuous spaces and/or the use of different approximations, most notably sampling algorithms.
Sampling in particular has become prominent not only because of its manageable computational requirements but also because there is some evidence that humans' deviations from optimal Bayesian reasoning, i.e. fallacies such as the base-rate neglect \cite{sanborn2016bayesian},
may be explained by considering human reasoning as approximate inference with a limited number of samples \cite{vul2014one}.

Due to both of these reasons, the need for specialized models and the challenges with efficient inference, have so far made it difficult to employ BToM models in live systems that try to infer a user's mental state in real-time. Our proposed Switching approach should be considered as one additional possible way of enabling statisficing ToM models for artificial systems.

\section{Scenario: Navigating a Maze with Different Uncertainties}


In order to be able to evaluate different models in situations where different mental states should be considered on actual human behavior data, we first needed to collect such data. 
The scenario we chose needed to be easy to understand for participants and simple enough to allow exact inference in our models, as we did not want to have our results influenced by approximate inference methods.
Furthermore, the scenario needed to allow us to influence the mental states of participants.
Due to these reasons we chose a navigation task in a 2D maze like environment where we influenced the amount of information provided to the participants in different conditions.

\begin{figure}
    \centering
	\includegraphics[width=\textwidth]{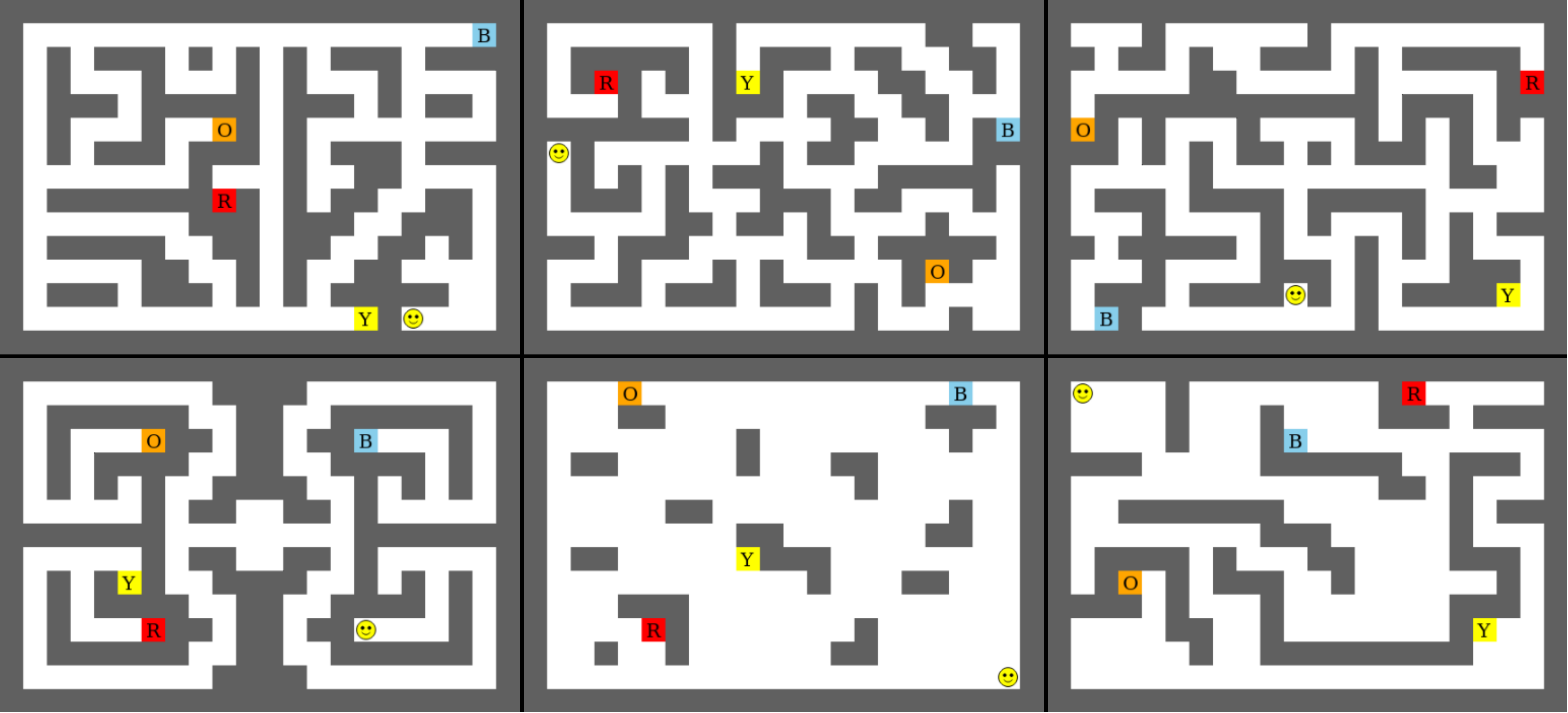}
	\caption{The different mazes (ordered 1-6 from top left to bottom right). 
	Colored squares represent potential exits and the smiley represents the agent to be navigated by participants, here shown in its starting position.}
	\label{fig:mazes}
\end{figure}

\begin{figure}
    \centering
	\includegraphics[width=0.45\textwidth]{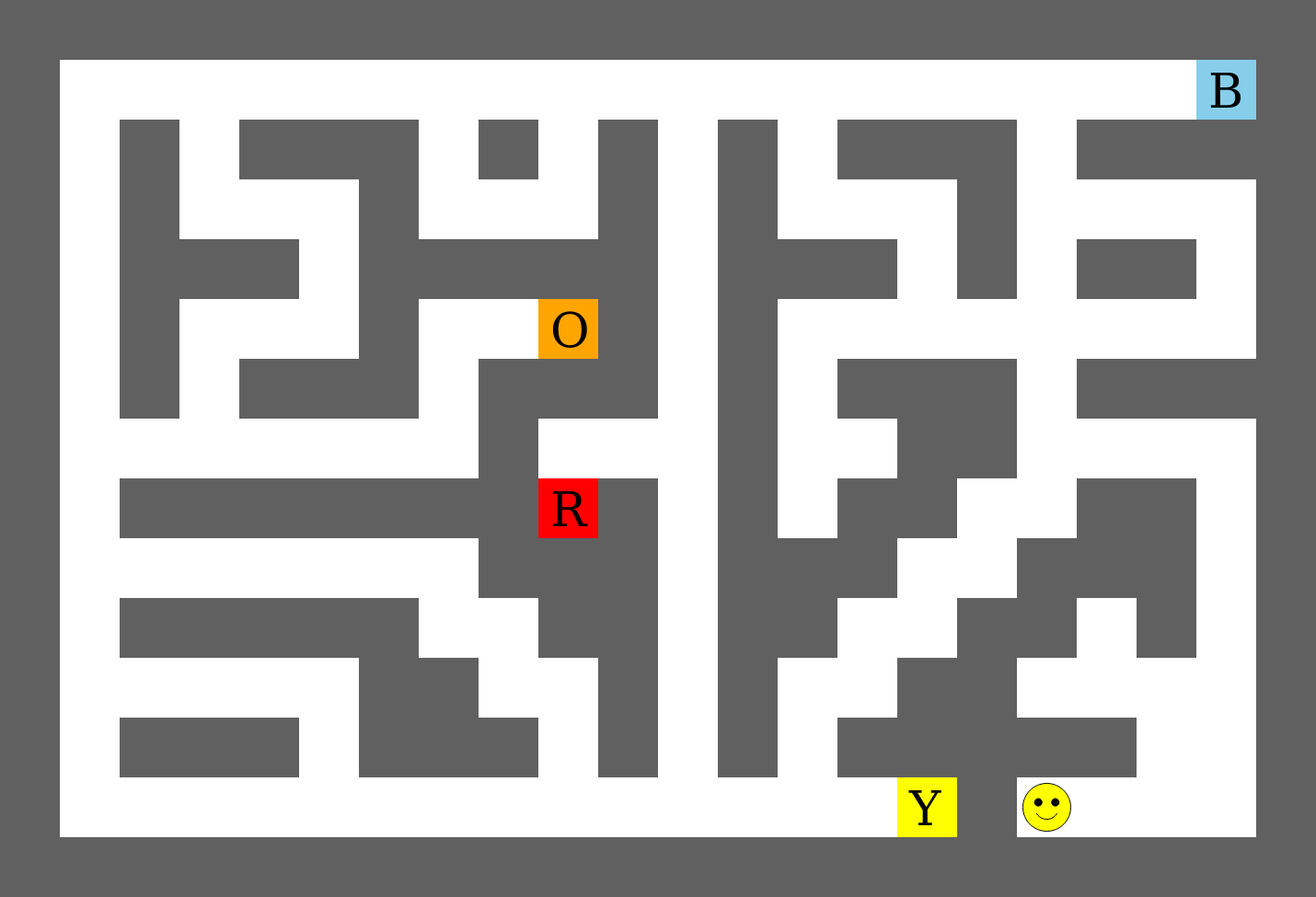}
	\includegraphics[width=0.45\textwidth]{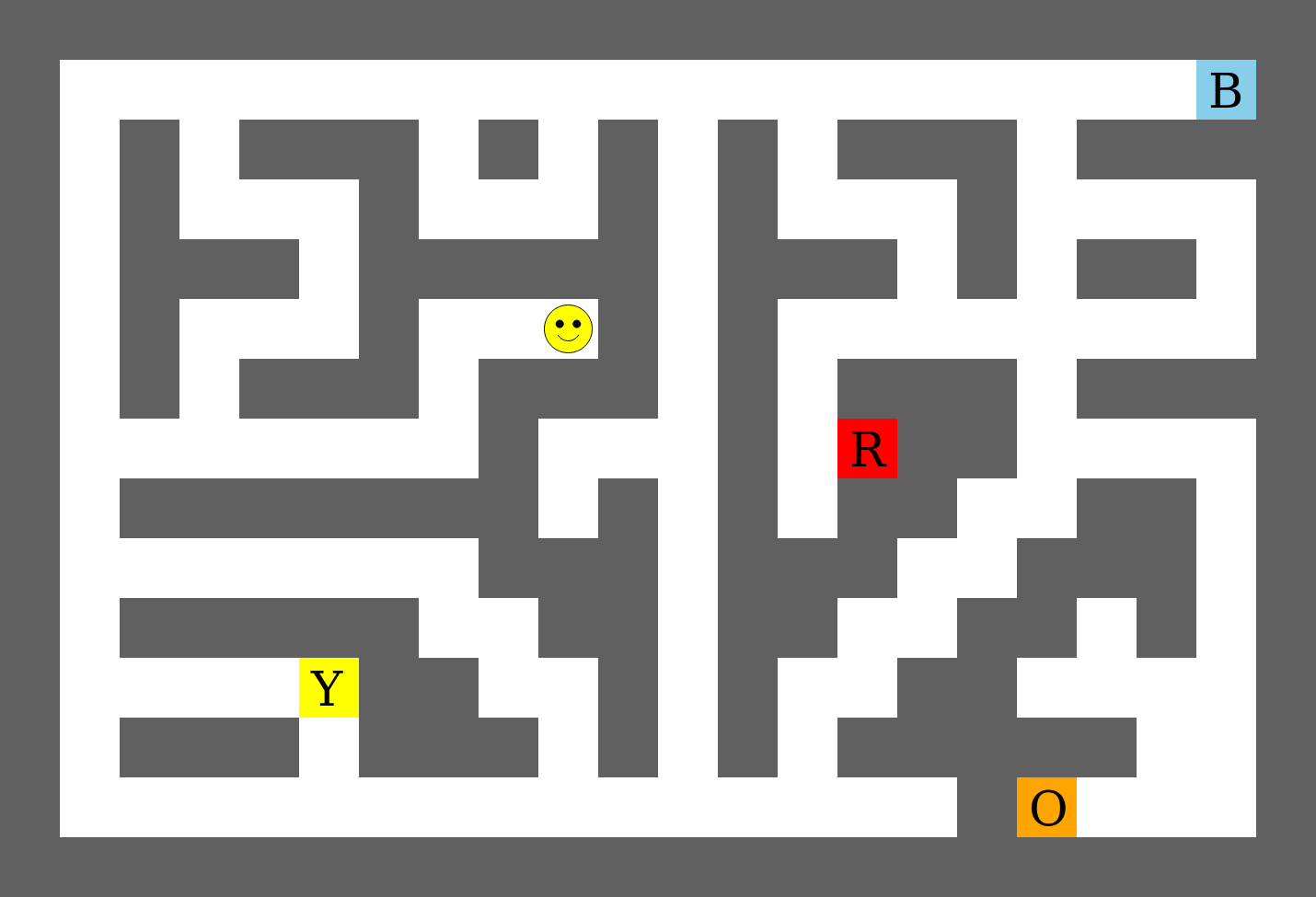}
	\caption{The two variants of Maze 1 used in the study. The smiley indicates the respective starting positions.} 
	\label{fig:variants}
\end{figure}

We chose such a ``gridworld'' environment because its discrete nature allows us to perform exact inference. On top of that, we can easily manipulate the amount of information we provide to participants and lastly, we reason that navigating an agent through simple environments is fairly intuitive to most participants.
We designed six different mazes of varying complexity which can be seen in Figure \ref{fig:mazes}. Each maze has four potential exits, differing by their color and corresponding symbol, (R)ed, (Blue), (Yellow) and (O)range.
For each maze we further introduced a second variation, where we changed the agent's starting position and/or the exit locations. An example of the two different variants of Maze 1 can be seen in Figure \ref{fig:variants}.

Similar gridworlds have also been successfully used in previous research using the BToM framework, such as \cite{Baker2009}, however in these cases the behavior that is later used as stimuli for their models and participants is usually handcrafted.

\subsection{Conditions and respective Uncertainties \label{sec:conditions}}

\begin{figure}
	\includegraphics[width=\textwidth]{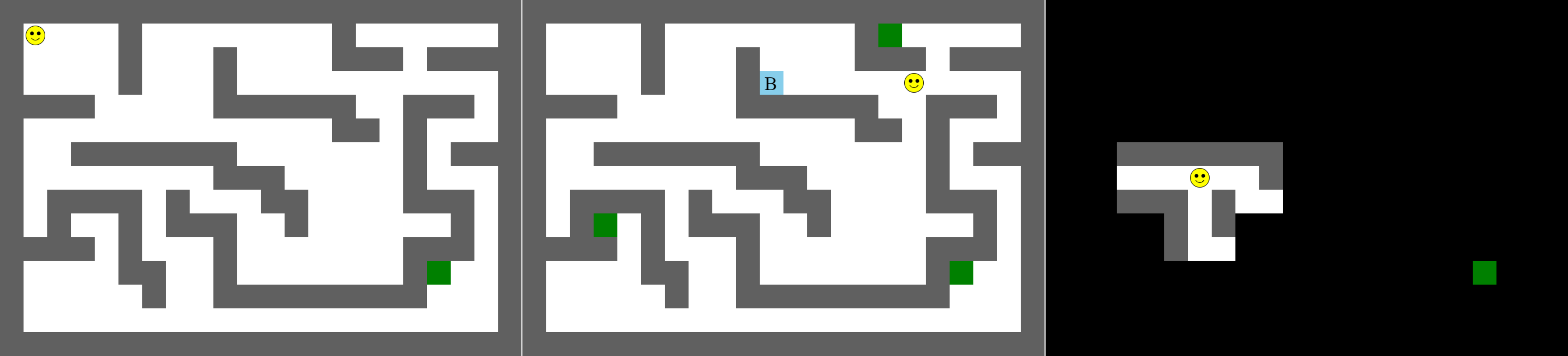}
	\caption{The three conditions \emph{No Uncertainty}, \emph{Destination Uncertainty} and \emph{Path Uncertainty}, respectively, from a participant's perspective, here for Maze 6.}
	\label{fig:conditions}
\end{figure}

In order to gather behavior data produced by different mental states
we modified the amount of information available to participants.
That is, we induced different amounts of uncertainty, which in turn can be interpreted as resulting in different mental states.
We used three different conditions differing in how the mazes were presented to participants as shown in Figure \ref{fig:conditions} for Maze 6.

\paragraph{No Uncertainty}

The first condition (leftmost image in Figure \ref{fig:conditions}) was termed \emph{No Uncertainty} (NU). 
In this condition, we removed all sources of uncertainty with respect to the navigation task by showing the entire maze with one exit to the participants. 
The exit is initially only shown in a generic green color. Its actual color is only revealed once a line of sight is established (compare with the middle image in Figure \ref{fig:conditions} of the \emph{Destination Uncertainty} condition).
As such we assume participants in this condition should hold a correct and true belief regarding the structure of the maze as well as their belief about where the exit they are supposed to reach, i.e. their goal, is located.

\paragraph{Destination Uncertainty}

In the second condition, termed \emph{Destination Uncertainty} (DU), we again show the entire maze, but this time include all four potential exits, each hiding their actual color/identity unless a line of sight is established between the exit and the agent (see center image in Figure \ref{fig:conditions}).
Participants may know the color of the desired exit, but were uncertain about which exit had which color initially.
Unlike in the first condition, participants could not have a certain a belief about the location of the desired exit in this case. 

\paragraph{Path Uncertainty}

The third condition, called \emph{Path Uncertainty} (PU), introduces uncertainty in 
the belief or knowledge regarding the structure of the maze.
In this condition participants got to see only a single exit, similar to the first condition, however, we only revealed up to three blocks around the agent, thus hiding the rest of the maze (see rightmost image in Figure \ref{fig:conditions}). That way participants could not be certain regarding the paths leading towards the exit.

\section{Collection of Empirical Behavior Data \label{sec:behaviorData}}

In order to collect human behavior data generated by different mental states, we conducted an online study using the Crowdflower platform (now re-branded as figure-eight\footnote{\url{https://www.figure-eight.com/}}). We recruited 122 participants via this service, which were asked to complete our web-based survey in which they had to navigate through all six mazes, after which they received 0.2\euro.
Out of the 122 participants, only 110 completed all six mazes. The remaining 12 stopped after 1 to 5 mazes.

In total we collected 687 complete behavior trajectories
that reached to the actual exit within the different maps and conditions as shown in Table \ref{tab:completions}.

\subsection{Procedure}

All participants were asked to navigate all six mazes with each condition appearing twice. 
Which maze was presented in which condition (and variant) was randomly assigned to the participants, only ensuring that each participant encountered each maze only once and each condition twice.

Participants were initially instructed that they would need to navigate a number of small mazes to reach a specific exit. 
Participants were then shown the maze, informed that they could control the agent in the four cardinal directions either using their keyboard or the buttons located around the maze and were given a condition-specific task instruction:
In the \emph{No Uncertainty} condition participants were told ``Reach the shown exit.''.
In the \emph{Destination Uncertainty} condition the instructions read ``Find the <color> (<colorsymbol>) exit.'', where the color of the actual target and its symbol as rendered on the maze ((R)ed, (B)lue, (O)range or (Y)ellow) was inserted accordingly. 
Finally, in the \emph{Path Uncertainty} condition participants were told ``Find your way to the shown exit.''

We also told participants that they needed to specifically ``enter'' (either using the E or the Return key on their keyboard or the corresponding button on the website) the exit in order to complete a maze. Using enter on any tile other than the correct exit did nothing.

The instructions were initially shown in an overlay on top of the maze, which participants had to dismiss by pressing a key. 
Furthermore, the task description as well as all control instructions were always present beneath the maze so that participants could consult them as needed.

As soon as a participant finished a maze, the next maze would load and showing the corresponding task instructions.
After finishing all six mazes, participants were given a password, which they could use to complete the survey on the Crowdflower platform.

\subsection{Results}

As we will use this behavior data for evaluating computational models (see Section \ref{sec:models}), we present here a brief summary of the data.

\paragraph{Completions}
Since we assigned conditions and variants randomly and since not all participants completed all 6 mazes, we ended up with different numbers of completed trajectories for each of the different variants. Table \ref{tab:completions} summarizes these completed trajectories for all mazes, conditions and variants.

\pgfplotstableread[col sep=tab,]{data/completions_table.csv}\datatable
\begin{table}
	\centering
	\pgfplotstabletypeset[
		column type=c,
		every head row/.style={
			before row={
				\toprule
				& \multicolumn{2}{c}{NU} & \multicolumn{2}{c}{DU} & \multicolumn{2}{c}{PU}\\
			},
			after row=\midrule,
		},
		every last row/.style={
			after row=\bottomrule},
		every first column/.style={
			column type/.add={}{|}
		},
		columns/map/.style  ={column name=},
		columns/C1V1/.style  ={column name=V1},
		columns/C1V2/.style ={column name=V2},
		columns/C2V1/.style  ={column name=V1},
		columns/C2V2/.style={column name=V2},
		columns/C3V1/.style  ={column name=V1},
		columns/C3V2/.style={column name=V2},
		col sep=&,row sep=\\,string type,]{\datatable}
	\caption{Breakdown of the 687 completed trajectories for the different mazes, conditions and variants. NU, DU and PU are the different conditions \emph{No Uncertainty}, \emph{Destination Uncertainty} and \emph{Path Uncertainty} respectively, cf. Section \ref{sec:conditions}}
	\label{tab:completions} %
\end{table}

We can see that most variants were completed somewhere between 15 and 28 times, with a strong outlier being the first variant for the Path Uncertainty condition in maze 6, where we unfortunately only got 7 valid trajectories.

\paragraph{Average path lengths}

Since we are assuming that our participants act like rational agents, we expected them to solve the task of reaching their goal close to optimal. Table \ref{tab:optimality} presents the optimal number of steps required to reach the goal from the start position as well as how many more steps participants required on average for the different conditions and variants.

\pgfplotstableread[col sep=tab,]{data/optimality.csv}\datatable
\begin{table}
	\centering
	\pgfplotstabletypeset[
	column type=c,
	every head row/.style={
		before row={
			\toprule
			& \multicolumn{2}{c|}{Optimal} & \multicolumn{2}{c}{NU} & \multicolumn{2}{c}{DU} & \multicolumn{2}{c}{PU}\\
		},
		after row=\midrule,
	},
	every last row/.style={
		after row=\bottomrule},
	every first column/.style={
		column type/.add={}{|}
	},
	columns/map/.style ={column name=},
	columns/optV1/.style ={column name=V1},
	columns/optV2/.style ={column type=c|, column name=V2},
	columns/C1V1/.style  ={column name=V1,
	    postproc cell content/.append style={/pgfplots/table/@cell content/.add={}{\%}},
	},
	columns/C1V2/.style={column name=V2,
	    postproc cell content/.append style={/pgfplots/table/@cell content/.add={}{\%}},
	},
	columns/C2V1/.style  ={column name=V1,
	    postproc cell content/.append style={/pgfplots/table/@cell content/.add={}{\%}},
	},
	columns/C2V2/.style={column name=V2, 
	    postproc cell content/.append style={/pgfplots/table/@cell content/.add={}{\%}},
	},
	columns/C3V1/.style  ={column name=V1,
	    postproc cell content/.append style={/pgfplots/table/@cell content/.add={}{\%}},
	},
	columns/C3V2/.style={column name=V2,
	    postproc cell content/.append style={/pgfplots/table/@cell content/.add={}{\%}},
	},
	col sep=&,row sep=\\,string type,]{\datatable}
	\caption{Minimum number of steps required for the different mazes and their variants as well as the percentage of the average increase in actual steps taken by the participants for each of the different mazes, their variants and conditions.
	}
	\label{tab:optimality}
\end{table} 

While the number of steps varies greatly between the different mazes, due to different lengths of the required paths, we see a clear trend: 
As expected, participants were most optimal in the \emph{No Uncertainty} condition, where only a small number of unnecessary actions were taken. 
Participants took substantially more steps in the \emph{Path Uncertainty} condition 
with the exception of both variants 
of Maze 5 and Variant 1 of Maze 2. All of these three variants have in common that a greedy strategy towards the goal corresponds closely to the optimal path.
While not as clear, participants generally took the most steps in the \emph{Destination Uncertainty} condition, where they did not know about the location of their goal.

\begin{figure}
    \centering
    \begin{subfigure}[t]{0.32\textwidth}
        \centering
        \includegraphics[width=\textwidth]{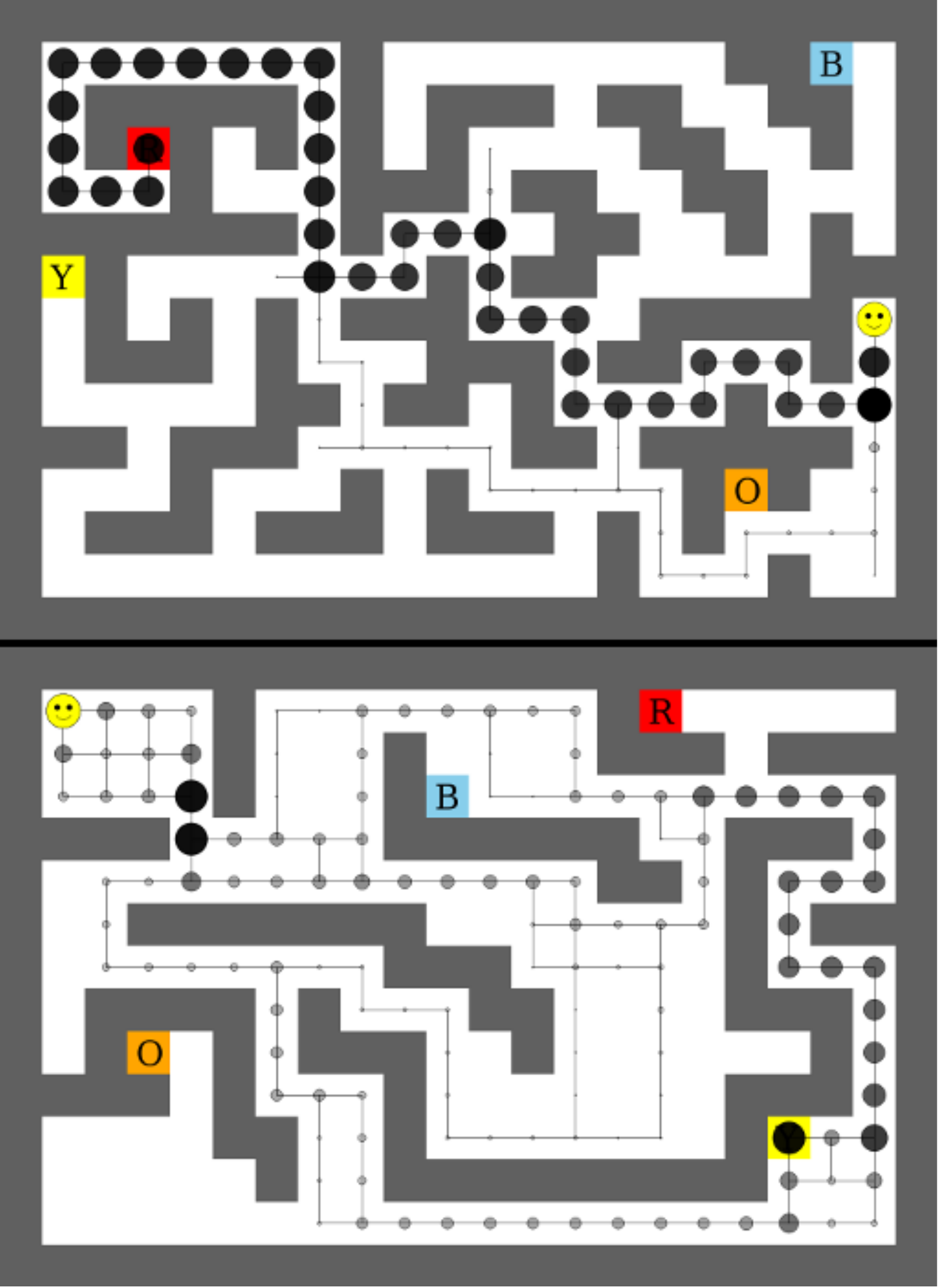}
        \caption{NU}
        \label{fig:aggregatesC1}
    \end{subfigure}
    \begin{subfigure}[t]{0.32\textwidth}
        \centering
        \includegraphics[width=\textwidth]{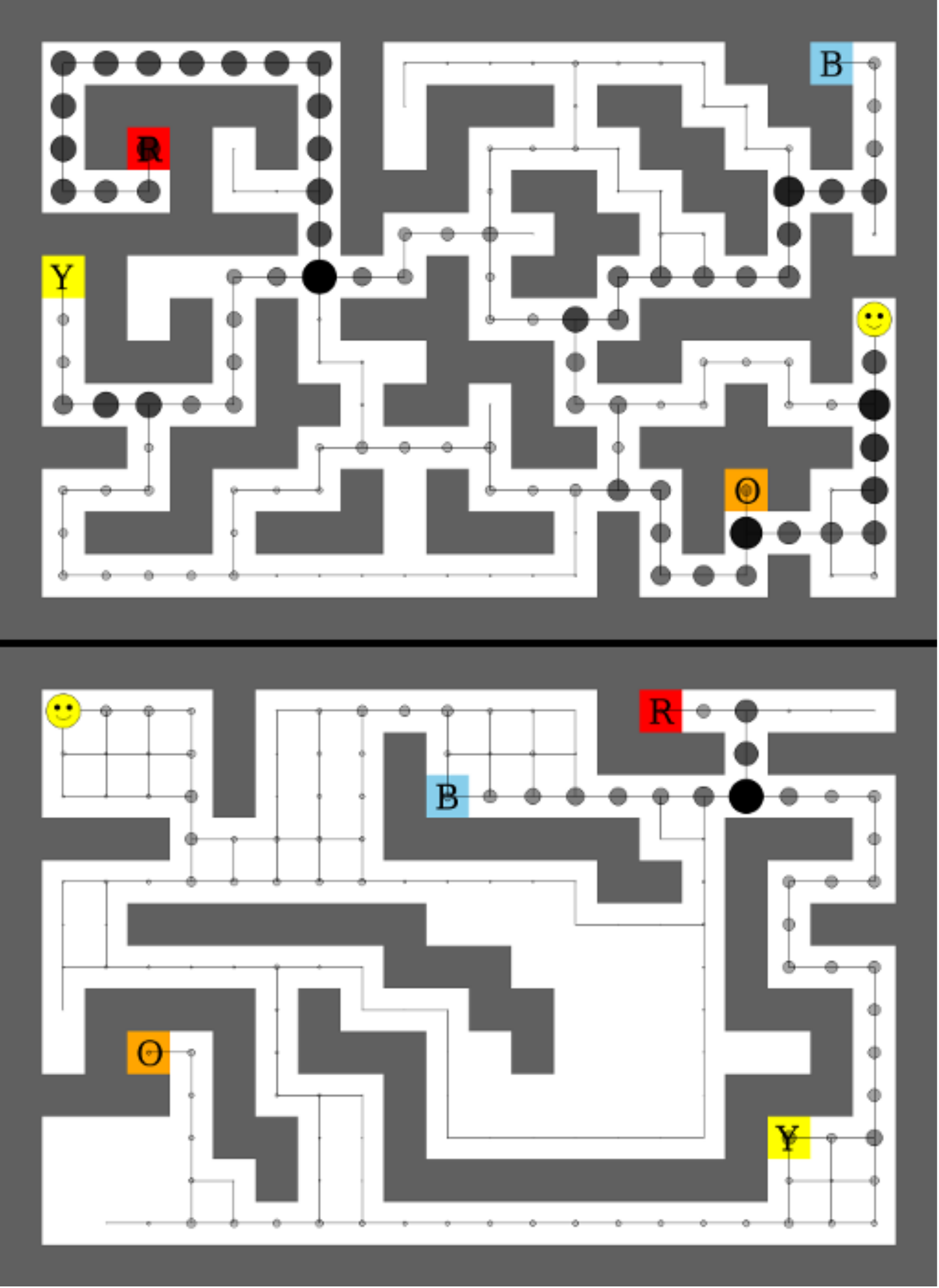}
        \caption{DU}
        \label{fig:aggregatesC2}
    \end{subfigure}
    \begin{subfigure}[t]{0.32\textwidth}
        \centering
        \includegraphics[width=\textwidth]{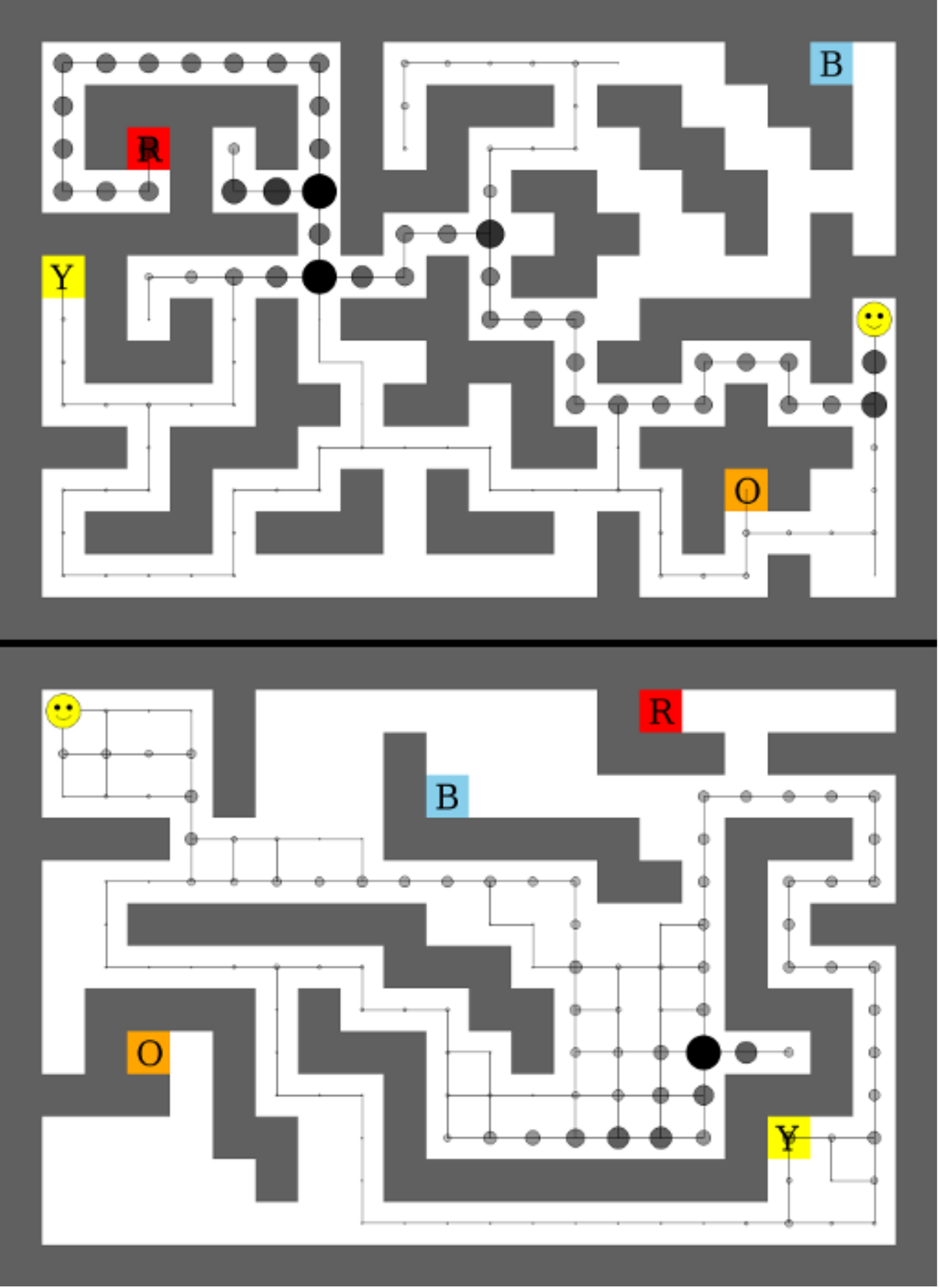}
        \caption{PU}
        \label{fig:aggregatesC3}
    \end{subfigure}
	\caption{Aggregation of all trajectories in Maze 2, Variant 2 for the top row (the goal was the red exit) and maze 6, variant 1 for the bottom row (the goal was the yellow exit) as black lines. 
	Trajectories from the three conditions \emph{No Uncertainty}, \emph{Destination Uncertainty} and \emph{Path Uncertainty} are shown in (a), (b) and (c) respectively. Circle size and opacity represents the relative number of times a cell was visited. The smiley represent the starting position in each case.}
	\label{fig:aggregates}
\end{figure}

The differences in behavior between the three conditions can be seen more clearly if one considers the percentage of trajectories that performed (close to) optimal, i.e. trajectories where participants used (close to) the minimum number of steps to reach their goal. 
Note, that rationally the different conditions would have different ``optimal'' behavior. Here, we only consider optimality from the point of view of an oracle in order to highlight that the different conditions successfully evoked different behavior.
Figure \ref{fig:optimalitiesConditions} shows the percentages across all mazes and variants for the three conditions, while distinguishing between optimal trajectories and those that took 10\% or 20\% additional steps. 
We include these 10\% and 20\% evaluations since with this we can still count participant that only overshot at an intersection (see Figure \ref{fig:aggregates} for examples) as behaving close to optimal as well\footnote{Note that due to the nature of the mazes, any erroneous action will need to be corrected, meaning that for each error, the participant will increase the number of steps taken by two.}.
The plot reveals clear differences between the three conditions. 
While in the \emph{No Uncertainty} condition over 83\% of trajectories were completed with a 20\% margin around the optimal number of steps, in the \emph{Destination Uncertainty} condition less then 8\% of participants took close to the minimum number of steps, as one would expect, since participants needed to search for their goal exit. 
The \emph{Path Uncertainty} condition is in the middle, with close to 32\% of trajectories being completed near optimally. 
However, it is noteworthy that the fraction of trajectories actually being completed optimally is a lot lower in the \emph{Path Uncertainty} condition compared to both other conditions. Furthermore, this fraction is a lot more dependent on the maze compared to the \emph{Destination Uncertainty} condition, as can be seen in Table \ref{tab:optimality} and as well as the examples in the next section.

This general trend can be seen across all mazes, although there are differences between the different mazes as well as even just variants of the same maze. A fairly extreme example of differences between variants is Maze 2 as shown in Figure \ref{fig:optimalitiesMaze2}.
While the general points still hold, we see almost similar percentages of almost optimal behavior in the \emph{Path Uncertainty} condition as in the \emph{No Uncertainty} condition for Variant 1. 
However, this chances in the second variant, where not even half as many participants performed the task nearly optimal in the \emph{Path Uncertainty} condition.

\pgfplotstableread[col sep=tab,]{data/optimalitiesByCondition.csv}\datacsv
\begin{figure}
    \centering
	\begin{tikzpicture}
	\begin{axis}[
	ybar stacked, 
	ymin=0, ylabel=Percentage,
	xlabel=Condition,
	xtick=data,
	xticklabels={NU, DU, PU},
    grid=major,
    xmajorgrids=false,
    legend style={
            area legend,
            at={(0.5,1.1)},
            anchor=center,
            legend columns=-1,
        }
	]
	\addplot table [y=optimal]{\datacsv};
	\addplot table [y=optimal10]{\datacsv};
	\addplot table [y=optimal20]{\datacsv};
	\addplot table [y=others]{\datacsv};
	\legend{Optimal, Optimal+10\%, Optimal+20\%, Other}
	\end{axis}
	\end{tikzpicture}
	\caption{Percentage of trajectories that were (close to) optimal for the three different uncertainty conditions. Optimal+10\% and Optimal+20\% refer to trajectories that used 10\% or 20\% more steps than the optimum respectively. All other trajectories with more steps are collected under Other. }
	\label{fig:optimalitiesConditions}
\end{figure}
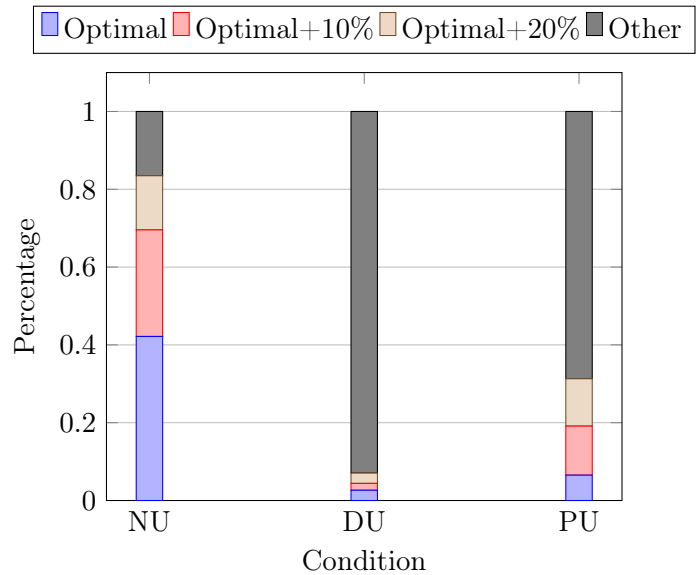

\begin{figure}
	\centering
	\pgfplotstableread[col sep=tab,]{data/optimalities_2.csv}\datacsv
	\begin{subfigure}[t]{\textwidth}
		\centering
		\ref{leg:optimalM2}
	\end{subfigure}
	\begin{subfigure}[]{0.26\textwidth}
	    \centering
		\raisebox{0mm}{\includegraphics[width=\textwidth]{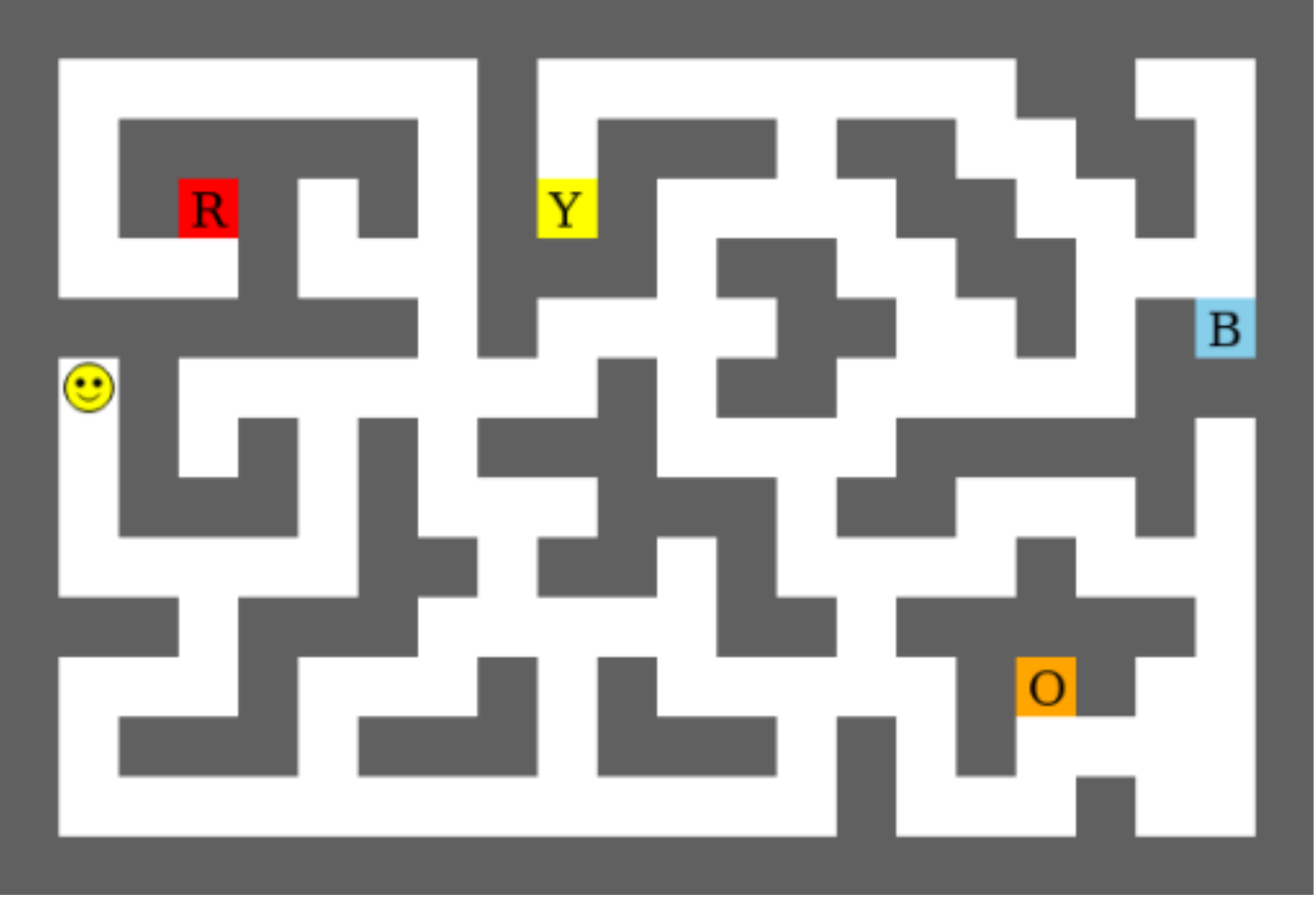}}
		Variant 1
	\end{subfigure}
	\begin{subfigure}[]{0.45\textwidth}
		\begin{tikzpicture}
		\node [align=center,
		text width=3cm, inner sep=0.25cm] at (0.75cm, -0.7cm) {\textsc{NU}};
		\node [align=center,
		text width=3cm, inner sep=0.25cm] at (2.7cm, -0.7cm) {\textsc{DU}};
		\node [align=center,
		text width=3cm, inner sep=0.25cm] at (4.5cm, -0.7cm) {\textsc{PU}};
		\begin{axis}[
		width=\textwidth,
		height=0.2\textheight,
		ybar stacked, 
		ymin=0, 
		xtick=data,
		xticklabels={V1, V2, V1, V2, V1, V2},
		ytick={0,0.25,0.5,0.75,1},
		yticklabels={0, 25\%, 50\%, 75\%, 100\%},
		grid=both,
		legend columns=-1,
		legend entries={Optimal, Optimal+10\%, Optimal+20\%, Other},
		legend to name=leg:optimalM2,
		]
		\pgfplotsinvokeforeach {optimal,optimal10, optimal20, others}{
			\addplot table [x expr={\coordindex-mod(\coordindex,2)/4},y=#1]{\datacsv};
		}
		\draw (axis cs:1.375,0) -- ({axis cs:1.375,0}|-{rel axis cs:0.5,1});
		\draw (axis cs:3.375,0) -- ({axis cs:3.375,0}|-{rel axis cs:0.5,1});
		\end{axis}
		\end{tikzpicture}
	\end{subfigure}
	\begin{subfigure}[]{0.26\textwidth}
	    \centering
		\raisebox{0mm}{\includegraphics[width=\textwidth]{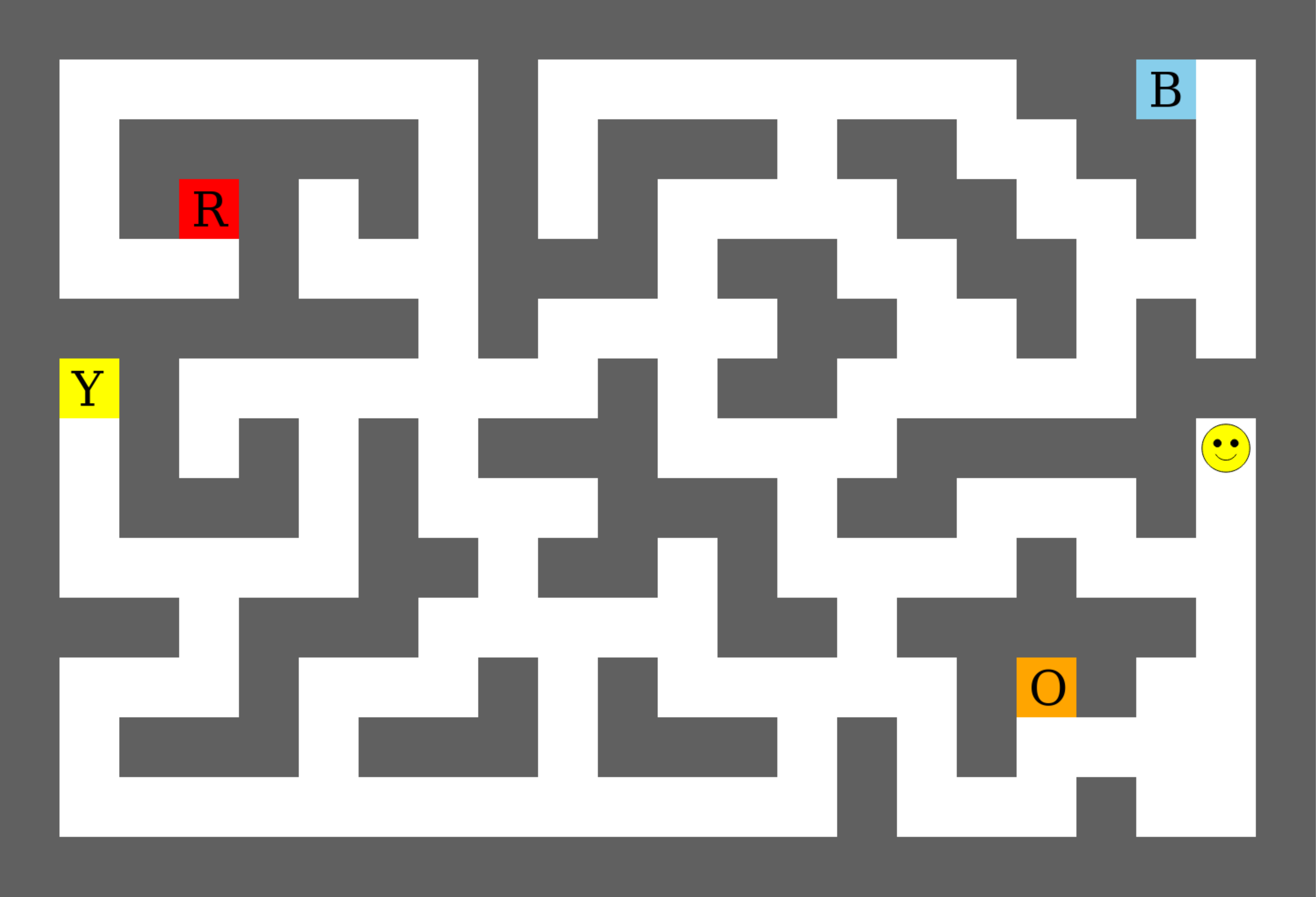}}
		Variant 2
	\end{subfigure}
	
	\caption{Center: Percentage of trajectories that were (close to) optimal for the three different uncertainty (U) conditions. Optimal+10\% and Optimal+20\% refer to trajectories that used 10\% or 20\% more steps than the optimum respectively for both variants of Maze 2. All other trajectories with more steps are collected under Other. In the first variant (left image) the goal was the (B)lue exit, while it was the (R)ed exit in the second variant (right image).}
	\label{fig:optimalitiesMaze2}
\end{figure}
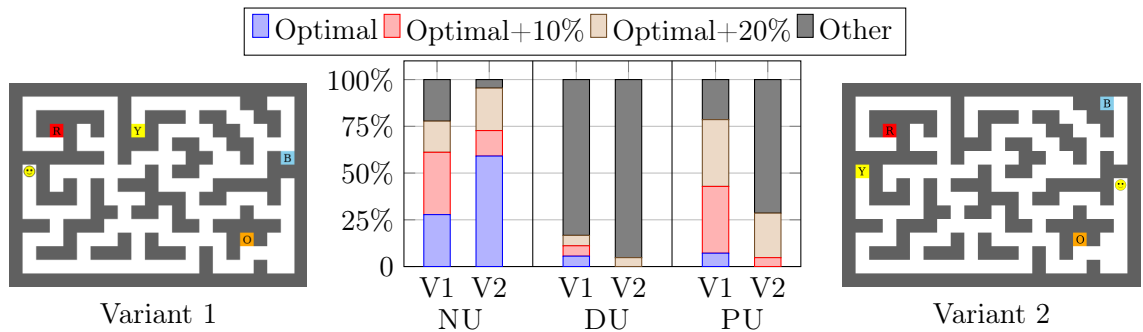

\paragraph{Visualization of the different Behaviors}

In order to visualize the differences in behavior, Figure \ref{fig:aggregates} shows aggregations from all trajectories of the \emph{No Uncertainty} (Figure \ref{fig:aggregatesC1}), \emph{Destination Uncertainty} (Figure \ref{fig:aggregatesC2}) and \emph{Path Uncertainty} (Figure \ref{fig:aggregatesC3}) conditions in Maze 2, Variant 2 (top row) and Maze 6, Variant 1 (bottom row). 
One can see clearly how most people follow one of the possible optimal ways in the \emph{No Uncertainty} condition (leftmost images), while they are employing a nearest-next exploration strategy in the \emph{Destination Uncertainty} condition (images in the middle). 
The difference in behavior between the \emph{No Uncertainty} and \emph{Path Uncertainty} conditions is less clear in Maze 2 (outer images in the top row) than in Maze 6 (bottom row) as a greedy homing strategy, which participants usually employed when they did not see the maze, is very close to the optimal trajectory in Maze 2.
Only the number of participants overshooting the last 4-way intersection and exploring the dead end pointing towards the red exit clearly hints at a difference in condition. 
On the other hand, in Maze 6 the homing strategy leads participants towards the dead end, that almost all participants run into in the \emph{Path Uncertainty} condition.


\section{Model Formulations\label{sec:models}}

The previous section revealed mostly different behavior depending on the knowledge or the amount of uncertainty participants are exposed to.
Next, we want to test ToM models which observe the generated data, try to infer the behaving agents' underlying mental states and use these to predict the agents' future actions.
To this end, we will define different Bayesian models of ToM reasoning and compare them across these different situations. 
We will first present a \emph{Full BToM} model designed to take into account all three kinds of uncertainty. We will then show how to simplify this model in order to end up with specialized models, each geared to one of the conditions.
Finally, we will describe a strategy for satisficing mentalizing that switches between these specialized models.

\subsection{The Full BToM model}

The three conditions in our scenario control the amount of uncertainty within the participants. We model these uncertainties using two different kinds of (discretized) beliefs, which we term world belief $B_w$ and goal beliefs $B_g$. 

Following the classical BToM framework, we need to define a generative model which incorporates these two beliefs, along with an agent's desire or \emph{goal} to determine the agent's next action $a_{t+1}$:

\begin{equation}
\label{eq:fullModel}
P(a_{t+1}|\boldsymbol{a}_{t}) = \sum_{\substack{g \in G \\ b_{g} \in B_{g} \\ b_w \in B_{w}}} P(a_{t+1}|g, b_{g}, b_{w}, \boldsymbol{a}_{t})  P(b_w| \boldsymbol{a}_{t})  P(b_g| \boldsymbol{a}_{t})  P(g|\boldsymbol{a}_{t}) 
\end{equation}

Here, $\boldsymbol{a}_{t}$ stands for the actions the agent performed in the previous $t$ time steps. We use $G$ as the set of possible goals the agent may have as its desire, i.e. the exit it wants to reach. 
The goal belief $B_g$ determines where one assumes a particular exit (and thus one's goal) to be located out of the four possible positions. 
In this scenario, the 4 exits can be arranged in a total of 24 different configurations.
Following \cite{Baker2011}, we split our goal belief $B_g$ into 24 possibilities, each representing one of these configurations. 

The world belief $B_w$ represents the agent's (believed) knowledge about the structure of the maze, i.e. the location of walls.
If we tried to tread $B_w$ the same as $B_g$ by considering all possible configurations of walls, even in our simple 2D environment of size 13 by 20, we would need to consider $2^{13*20}=1.85 \times 10^{78}$ different configurations. 
Even if one would a-priori select only worlds with reachable exits, one still ends up with an intractable number of possibilities. 
Since we do not seem it likely that humans are considering all potential wall placements either, we will employ a first heuristic for the \emph{Full BToM} model, the \emph{freespace assumption}.
The freespace assumption, commonly used in the robotic planning literature, e.g. \cite{koenig1997sensors,zelinsky1992mobile}, assumes that unknown cells are actually free, i.e. there is no wall and it is passable. 
Using the freespace assumption, we model $B_w$ as a binary variable, either the agent has a true belief regarding the maze's structure or the agent does not know anything about the maze until it was within its vision range and the agent uses the freespace assumption when considering distances. 


The likelihood $P(a_{t+1}|g, b_g, b_w, \boldsymbol{a}_t)$ is modeled similar to earlier ToM models, as that it models human actions according to the Boltzmann noisy rationality:

\begin{equation}
\label{eq:likelihood}
P(a_{t+1}|g, b_g, b_w) = \frac{exp(\beta U(a_{t+1}, b_g, b_w, g))}{\sum_{a_i \in A}exp(\beta U(a_i, b_g, b_w, g))}
\end{equation}

Here $\beta$ determines the degree of rationality. Higher values increase the probability to choose the action $a_{t+1}$ with the highest utility $U(a_{t+1}, b_g, b_w, g)$, whereas a values of $\beta=0$ would result in a uniform distribution across all possible actions.
The utility $U$ determines how beneficial the different actions are given the beliefs and the agent's goal. It follows from solving the Markov Decision Problem. By assuming equal costs for all actions, the utilities turn out to be equivalent to the remaining distance after executing action $a_{t+1}$ towards the exit determined by goal $g$ and the goal belief $b_g$ using either the true distance or the freespace assumption depending on $b_w$. 

$P(b_w| \boldsymbol{a}_{t})$, $P(b_g| \boldsymbol{a}_{t})$ and $P(g|\boldsymbol{a}_{t})$ are the prior probabilities for the world belief, the goal belief and the goal intention respectively. These are initially assumed to be uniform.
We simply use Bayes' rule to update $P(g|\boldsymbol{a}_{t})$ and $P(b_w| \boldsymbol{a}_{t})$ (compare the general BToM idea in \ref{eq:btom}):

\begin{align}
P(g|\boldsymbol{a}_{t}) &\propto P(\boldsymbol{a}_t| g) P(g) \\
\end{align}

and 

\begin{align}
P(b_w|\boldsymbol{a}_{t}) &\propto P(\boldsymbol{a}_t| b_w) P(b_w) \\
\end{align}

with prior probabilities $P(g)$ and $P(b_w)$. Note that these can also be computed incrementally, e.g. for the goal intention:

\begin{align}
P(g|\boldsymbol{a}_{t}) &\propto P(a_t|g) P(g|\boldsymbol{a}_{t-1})
\end{align}

However, we update our goal beliefs similar to \cite{Baker2011} by using an update rule based on observations:

\begin{equation}
P(b_g|\boldsymbol{a}_{t}) \propto P(o_t|s_t, b_g) P(b_g|\boldsymbol{a}_{t})
\end{equation}

Here we define $P(o_t|s_t,b_g)$\footnote{This is independent of the world belief $b_w$ as only visible exits are considered.} to be the probability of actually observing $o_t$ at time $t$ given the agent's current state $s_t$ and goal belief $b_g$. 
Since the goal belief only influences the believed exit locations, we compare if the assumed exit is visible from the agent's current position given its state. 
By assigning matches between the assumed and the actual observations the probability $\theta$, we can model noise sensors as well. 
However, in this work we do not focus on imperfect information, which is why we set $\theta$ to 1 for the results reported below.
Note that when no exit is visible from the agent's current position the observation matches the assumption.

\subsection{Specialized Models Based on Discrete Assumptions}

In contrast to the Full BToM model described above, we can also define specialized BToM models each tailored for the induced mental states in the three conditions we are considering here. 
Since the conditions only differ in the amount of uncertainty they introduce, the specialized models can actually be understood as making different kinds of discrete assumptions about the different belief states in the form of \emph{true beliefs} within the full model:

The first model, corresponding to the \emph{No Uncertainty} condition is termed \emph{True World and Goal} (TWG) model, as we are assuming a true belief regarding both the layout of the maze as well as the actual colors of the exits, i.e. the agent's goal:

\begin{equation}
\label{eq:twg}
P_{twg}(a_{t+1}|\boldsymbol{a}_{t}) = \sum_{g \in G} P(a_{t+1}|g, b_g^*, b_w^*) P(g|\boldsymbol{a}_{t})
\end{equation}

where $b_g^*$ and $b_w^*$ stand for the true goal belief and the true world belief respectively. By making these discrete assumptions, the model simplifies greatly compared to the full model in Equation \ref{eq:fullModel}, as we do not need to consider all combinations of possible beliefs anymore.

Consequently, the model specialized for the \emph{Destination Uncertainty} condition, named \emph{True World} (TW) model hereafter, only makes the true belief assumption for the world belief:

\begin{equation}
\label{eq:trueWorld}
P_{tw}(a_{t+1}|\boldsymbol{a}_{t}) = \sum_{\substack{g \in G \\ b_g \in B_g}} P(a_{t+1}|g, b_g, b_w^*) P(b_g|\boldsymbol{a}_{t}) P(g|\boldsymbol{a}_{t})
\end{equation}

Finally, the model for the \emph{Path Uncertainty} condition, called \emph{True Goal} (TG) model only makes this assumption for the goal belief:

\begin{equation}
\label{eq:trueGoal}
P_{tg}(a_{t+1}|\boldsymbol{a}_{t}) = \sum_{\substack{g \in G \\ b_w \in B_w}} P(a_{t+1}|g, b_g^*, b_w) P(b_w|\boldsymbol{a}_{t}) P(g|\boldsymbol{a}_{t})
\end{equation}

Due to our simplification with regard to the freespace assumption, Equation \ref{eq:trueGoal} simplifies further if we want to make it match the \emph{Path Uncertainty} condition, since in this case we will assume $b_w^'$ as following the freespace assumption instead of considering both possibilities:
\begin{equation}
\label{eq:trueGoalActual}
P_{tg}(a_{t+1}|\boldsymbol{a}_{t}) = \sum_{g \in G} P(a_{t+1}|g, b_g^*, b_w^') P(g|\boldsymbol{a}_{t})
\end{equation}

These specialized models represent only those uncertainties that were induced by the different conditions. As such we expect them to be quite good at inferring the induced mental states from their corresponding conditions, while being worse when inferring behavior generated in one of the other two conditions.

\subsection{Switching Strategy}

In order to leverage both the computational (and predictive) benefits of the specialized models as well as the flexibility to be able to deal with different scenarios, we propose here a very simple switching strategy, which will autonomously choose the specialized models which it deems most promising.

This strategy will effectively start with the simplest and most efficient \emph{TWG} model, basically exhibiting a strong egocentric tendency by projecting the observer's true belief onto the agent. 
Each observable action the agent performs is then evaluated with regard to how well it matches the current model's prediction. 
We consider unmatched observations as \emph{surprising}, with the degree of surprise depending on how unexpected the observations are.
The choice of matching function will change the behavior of the approach and different possibilities exist. In this paper we consider two different surprise measurements as matching function, which will be presented in Section \ref{sec:surprise}. 
If the observed behavior deviates too strongly, i.e. exceeds a dynamic threshold $\gamma$, from what the current model would expect, the switching model re-evaluates its current model choice and adapts the threshold.
In our first naive approach, this re-evaluation simply evaluates the past behavior of the this agent with all available models and switches to the one that can explain the behavior best. 
Note that the approach may stick with the current model in case no other model can outperform it. 
In order to avoid constant re-evaluations, the threshold $\gamma$ is increased by 50\% each time the model reconsiders. This will effectively make the Switching approach more tolerant to missmatching behavior with each re-evaluation. This behavior can obviously be adapted to ones current domain. If one instead desires a constant sensibility to missmatches, one could simply add a constant factor upon each reconsideration.
The general approach is summarized in Algorithm \ref{lst:switching}.

\begin{algorithm}
    \caption{Pseudo-code for the Switching strategy}
    \label{lst:switching}
  \begin{algorithmic}[1]
    \Require{Models M} 
    \Require{Initial threshold $\gamma$}
    \Require{Observations $O$}
    \Statex
     \State curModel $\gets$ selectSimplest(M) 
     \State accumulatedSurprise $\gets$ 0
     \State currentEpisode $\gets$ []
     \ForAll{action \textbf{in} $O$}
        \State currentEpisode.append(action)
        \State surprise $\gets$ curModel.computeSurprise(action)
        \State accumulatedSurprise $\gets$ accumulatedSurprise + surprise
        \If{accumulatedSurprise > $\gamma$}
            \State curModel $\gets$ chooseNextModel(M, currentEpisode)
            \State $\gamma$ $\gets$ $\gamma * 1.5$
        \EndIf
      \EndFor
  \end{algorithmic}
\end{algorithm}

\begin{figure}
    \centering
	\includegraphics[width=0.45\textwidth]{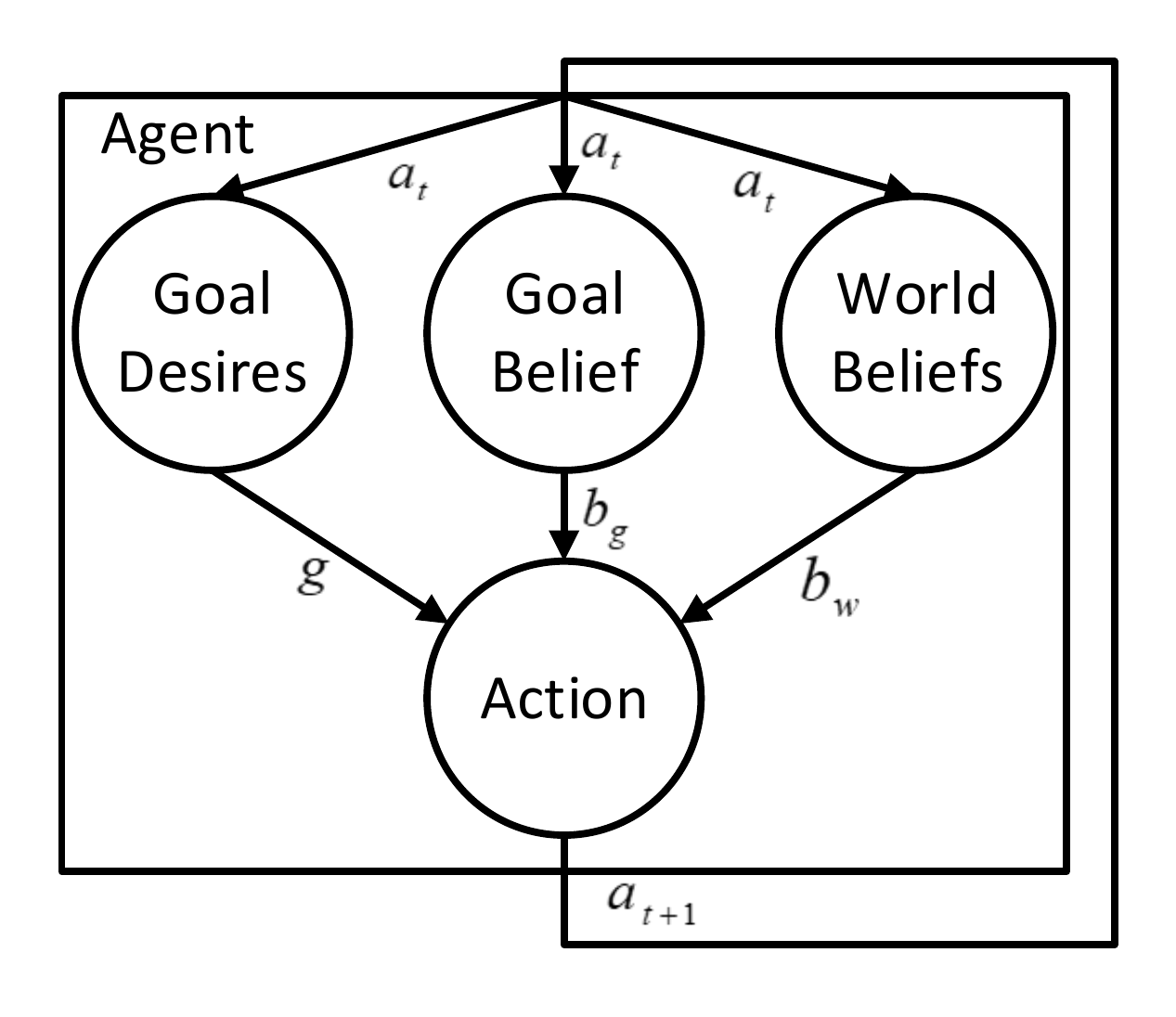}
	\includegraphics[width=0.45\textwidth]{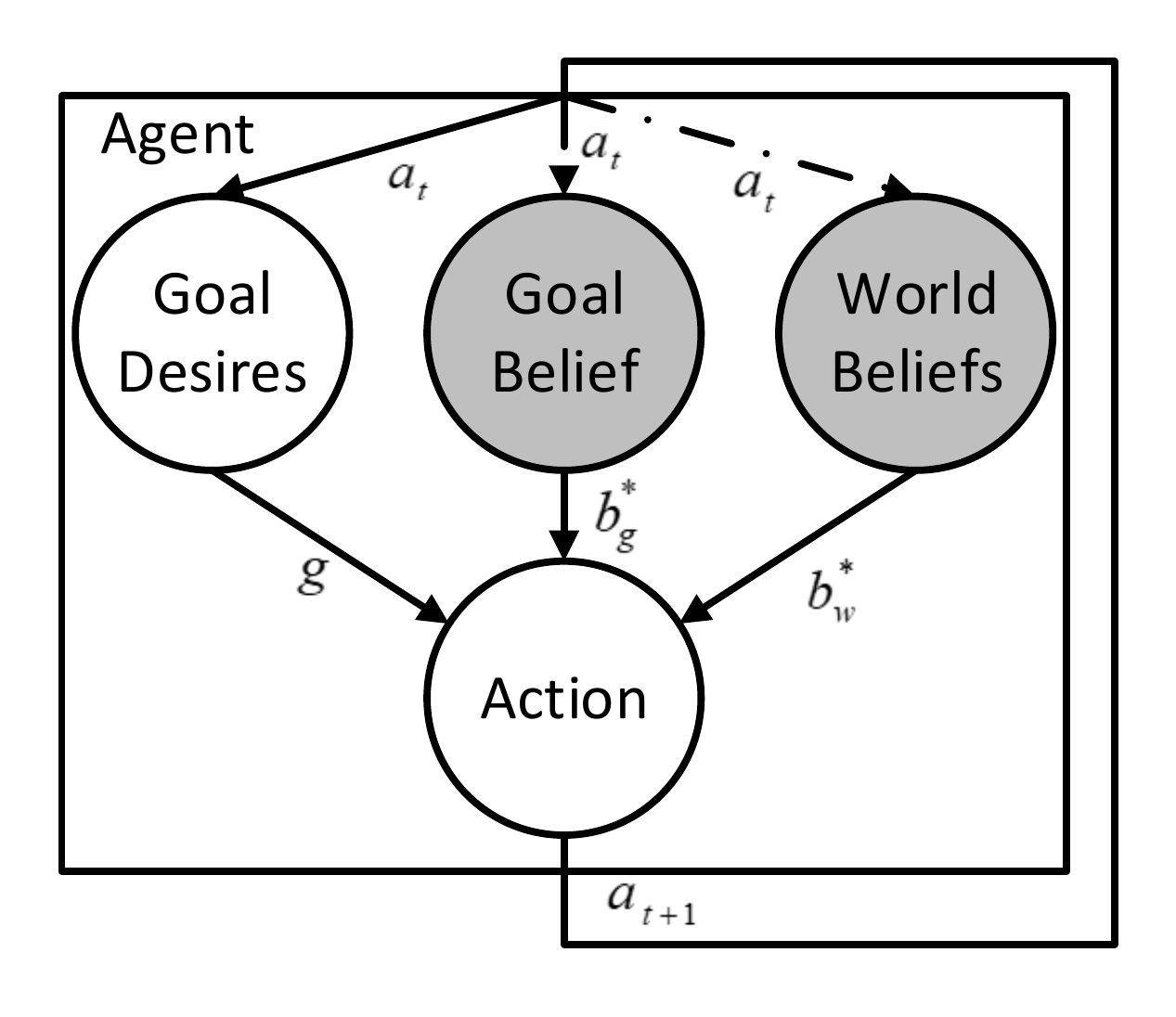} \\
	\includegraphics[width=0.45\textwidth]{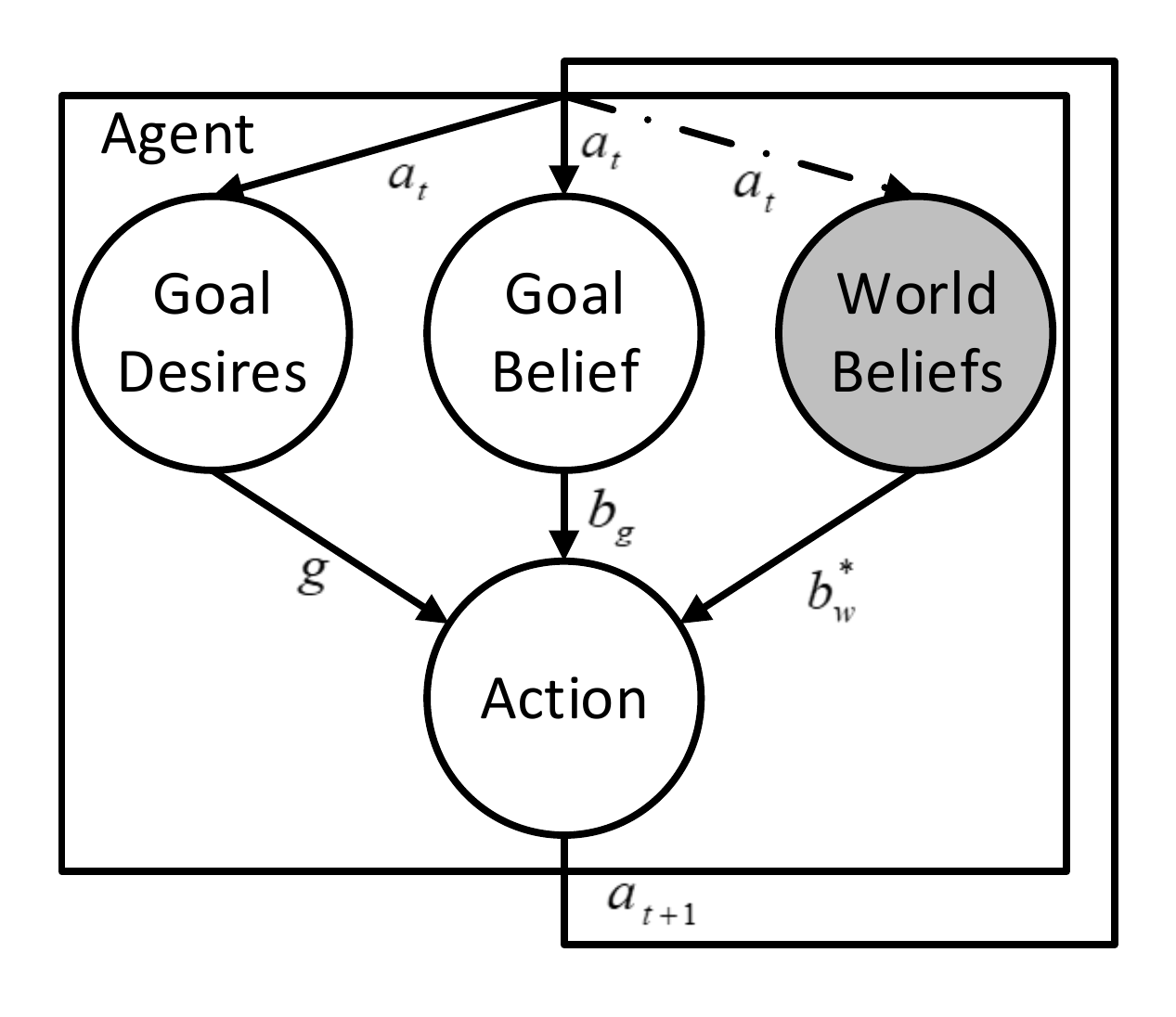}
	\includegraphics[width=0.45\textwidth]{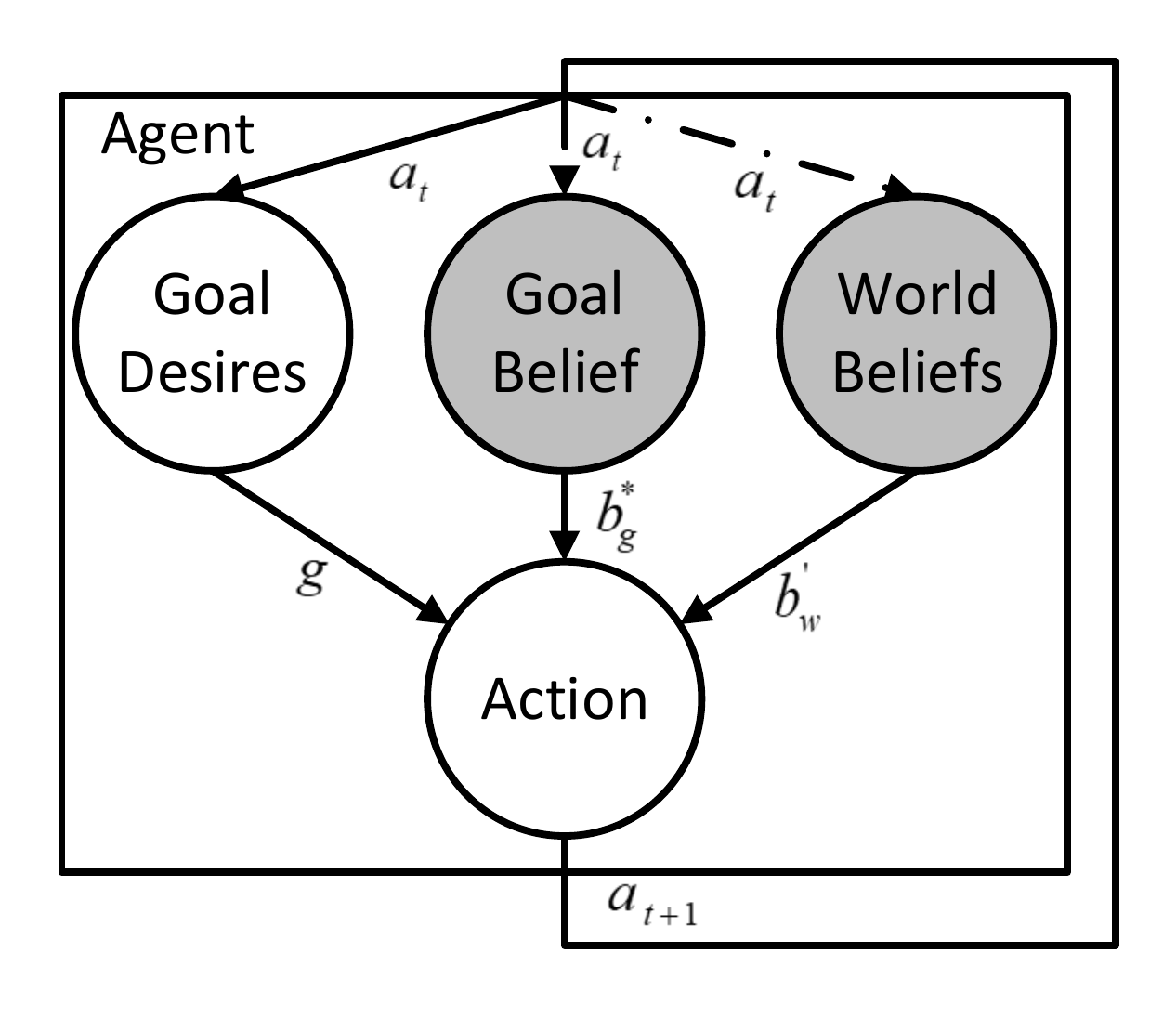}
	\caption{Visualization of the \emph{Full BToM} (top left), the \emph{TWG} (top right), the \emph{TW} (bottom left) and the \emph{TG} (bottom right) models. The circles represent random variables that are considered by the models. Grayed out circles are clamped to a certain belief (the true belief except for the TG model where it is $b_w^'$) so that not all possibilities are considered.}
	\label{fig:fullModel}
\end{figure}

\begin{figure}
    \centering
	\includegraphics[width=\textwidth]{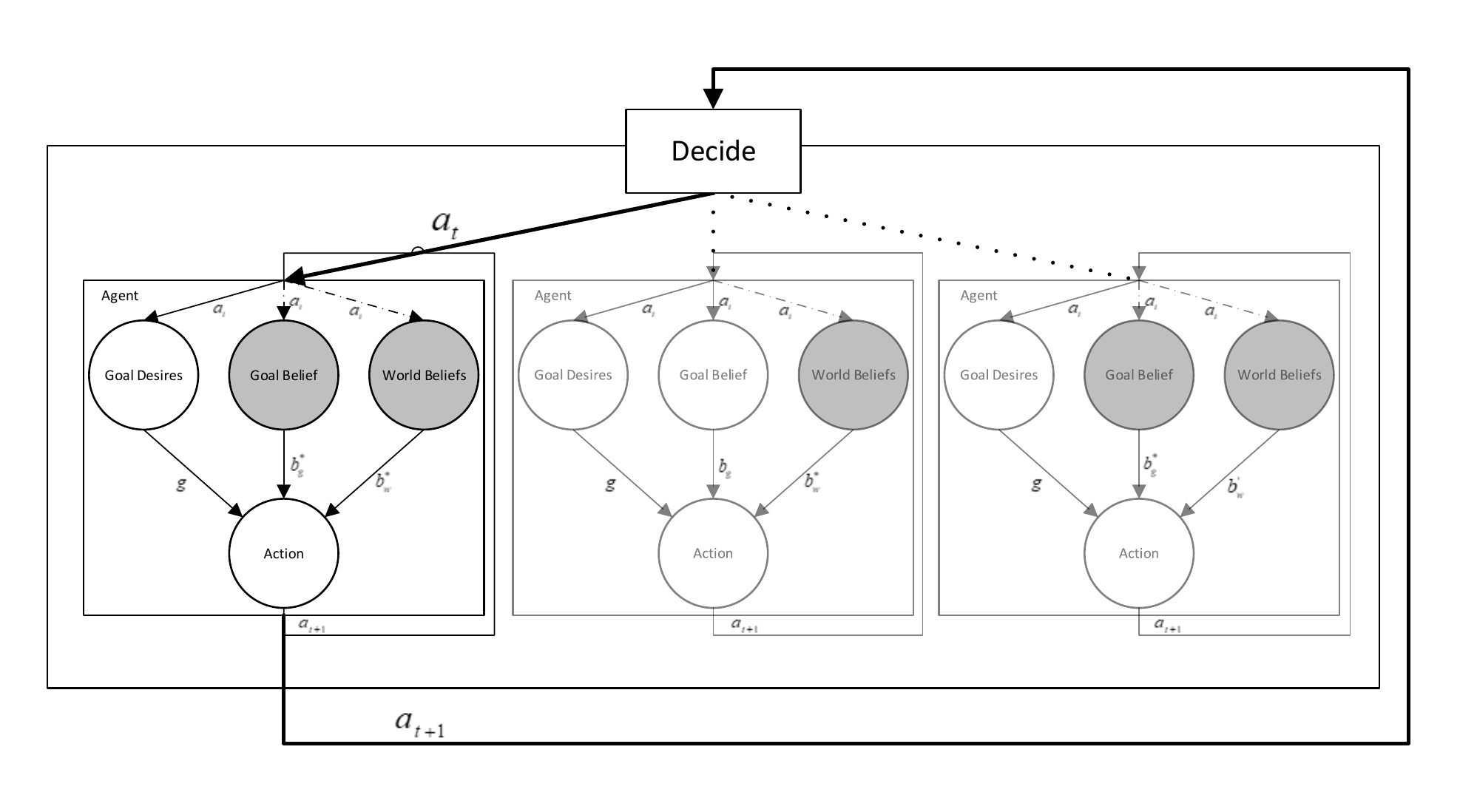}
	\caption{Visualization of the \emph{Switching} model. Only one of the specialized models is active at any time. The Decide process first determines whether or not the current model needs to be re-evaluated and if so will choose the next most suitable model.}
	\label{fig:SwitchingModel}
\end{figure}

All five different models are visualized in Figures \ref{fig:fullModel} and \ref{fig:SwitchingModel}. The (simplified) BToM models only really differ from each other in the number of mental states that they are actively evaluating. 
The Switching model can be considered as a meta-model, which uses the other simplified models adaptively. 

While we will present our results based on this very naive approach, we will discuss a range of promising modifications to this strategy in section \ref{sec:discussion} below.

\subsection{Surprise Measurements \label{sec:surprise}}

The switching approach outlined above requires us to evaluate an action or a sequence of actions regarding how well they match our models' predictions. 
As we mentioned above, we consider mismatches to be \emph{surprising}.
We chose this, as it has been found that humans tend to start a re-evaluation process upon being surprised \cite{meyer1997toward}, which is a process we imitate here.
However, it is not clear how to model surprise in artificial systems. There has been a range of different suggestions as to what a surprise measure could look like \cite{macedo2004modeling,itti2009bayesian}.

Arguably the simplest surprise measure when probabilistic action predictions are given is to just take the negative log-likelihood of the observed actions:

\begin{equation}
S_1(a) = -log(P(a))
\end{equation}

or when considering a sequence $\boldsymbol{a}_t = a_1,..,a_t$ of actions:

\begin{equation}
	S_1(a_1,..,a_t) = \sum_{i=1}^t -log(P(a_i)) 
\end{equation}

The log-likelihood is a common quality measure for probabilistic models and also known as the self-information in information theory.
Furthermore, it actually emerges naturally as a special case of the Kullback-Leibler (KL) divergence between the observation distribution $Q(a)$ and the predictive distribution $P(a)$, as long as one assumes perfect observation, such that $Q(a)=1$, iff a is the actually observed action:

\begin{equation}
S_{1}^'(a) = KL(Q(a)||P(a)) = \sum Q(a) \log(\frac{Q(a)}{P(a)})
\end{equation}

While this measure has many nice properties, it also comes with its own problems that limits its use as a surprise measure:
Since $S_1$ only looks at the self-information of the actually observed actions, it will punish actions that have equally likely alternatives. 
In any situation where there is more than one optimal action, $P(a)$ will at most be 0.5, resulting in fairly large surprise values from $S_1$ even though the action has been one of the optimal ones. This is especially detrimental in open environments such as Maze 5, where numerous equally good alternative paths towards any given goal exist.

In order to combat this trend, we consider a second surprise measure in this paper which has been used in \cite{macedo2004modeling} to explain human empirical data:

Instead of simply considering the log-likelihood, this measure considers the difference between the observed action's likelihood $P(a)$ and the probability of the most likely action according to the model $P_{max}$:

\begin{equation}
S_2(a) = log(1+P_{max}-P(a))
\end{equation}

This way, only actions that actually deviate from the predicted behavior are considered surprising. However, from a computational standpoint, this measure has the downside that it requires the computation of $P_{max}$ for each observed action, which is a lot more computationally expensive than the simple log-likelihood of $S_1$.

We like to point out that, regardless of surprise measurement used, a model's accumulated surprise can never go down when observing new actions, as an action will at most not produce any new surprise but never a negative surprise. 

\section{Results and Comparison of the Different Models} 

As introduced at the beginning of this paper, we are primarily interested in satisficing mentalizing. Satisficing, accoding to Simon \cite{simon1955behavioral}, is always a \emph{good enough} trade-off between optimality or accuracy and computational effort or time. 
What should be considered \emph{good enough} will change depending on any given situation.
Therefore, we will compare the different models regarding these two aspects making up satisficing systems, namely their predictive accuracy as well as their computational costs in terms of time. 
To analyse this, we have the models evaluate all of the 687 complete trajectories we collected in the study described in Section \ref{sec:behaviorData}.
Furthermore, in order to better highlight how the models, and the switching approach in particular, react to different human behavior, we will also present two example trajectories in more detail.


Finally, we have presented two possible alternatives for the surprise measurement used by the switching model. 
While these different measurements, change the way a model's predictions are interpreted and how ``surprising'' they are, we find that they make little differences with respect to the relative comparison between the different models. 
Therefore, in favor of better readability, we will primarily report our results using the negative log-likelihood/$S_1$ here, except for when the use of $S_2$ produces a qualitative difference to the model comparison. 
We used $\beta=1.5$ and an initial threshold of $\gamma=20$ 
for the \emph{Switching} model for all results reported here when using $S_1$ and $\gamma=1.5$ when using $S_2$.
A complete report of all data for both surprise measurements as well as on a more detailed scope can be found in \cite{poeppelJAIRSupplement} along access to the raw data and source code generating this data.

\subsection{Predictive Accuracy}

In order to determine the models' predictive accuracy, we present here the negative log-likelihood of the observed actions which corresponds to $S_1$. 
Lower values correspond to better predictions made by the models. 

Table \ref{tab:accuracyAggregated} summarizes the predictive performance of the different models in terms of their average negative log-likelihood aggregated across all different mazes and variants.

The table shows that the \emph{Switching} model outperforms all other models regardless of condition.

Furthermore, we see that the specialized models outperform the \emph{Full BToM} model in their respective conditions, except for the \emph{TW} model. 
However, as can be seen in Table \ref{tab:winningsTW}, when considering which model performed better more often in the \emph{Destination Uncertainty} condition, the \emph{TW} model comes out ahead of the \emph{Full BToM} model.


\accuracyTable{data/overallAccuracyKL_T20.csv}{tab:accuracyAggregated}{Average negative log-likelihood values (equivalent to $S_1$) for the different models aggregated over all trajectories for the different conditions and their standard deviations.}

We also notice fairly large standard deviations of the same order of magnitude of the mean. 
Considering the different mazes and their variants separately reveals strong differences between them. A complete breakdown of all 6 mazes with both variants can be found in \cite{poeppelJAIRSupplement} and Table \ref{tab:accuracy21} shows these results for Maze 2, Variant 1.

\accuracyTable{data/accuracyKL_T20_M2V1.csv}{tab:accuracy21}{Average surprise score according to $S_1$ for the different models aggregated over all trajectories in Maze 2, Variant 1.}

In this maze, the \emph{TG} model performs best in both the \emph{No Uncertainty} and the \emph{Path Uncertainty} conditions, even marginally better than the \emph{Switching} model.

\subsubsection{Pairwise Model Comparison}

Because of the fairly large standard deviations found in Table \ref{tab:accuracyAggregated} and the dependence of the maze as seen in Table \ref{tab:accuracy21}, 
we also looked at a head-to-head comparison of the different models. 
For each trajectory within our data, we compared each model against all other models, counting which achieved a lower or equal surprise value. Aggregating over the different mazes and/or conditions, this gives us the relative frequency of how often any particular model outperforms (or at least performs just as good as) another in that setting.
Table \ref{tab:winningsOverall} contains these results aggregated over all mazes and conditions, thus this is comparable to the \emph{Overall} column of Table \ref{tab:accuracyAggregated}. 
Using the $S_2$ surprise measurement, the overall picture remains the same. 
Only the \emph{Full BToM} and the \emph{TW} models benefit slightly from $S_2$, as considering different alternatives is not punished, as can be seen in Table \ref{tab:winningsOverallRel}.
The Switching model clearly outperforms any other model regardless of condition or surprise measure used. 
Note that the percentages of the Switching model winning and losing against other models may add up to more than 100\%. These include the cases where the Switching model performs the same as another model, as we are counting less or equal surprise values. The switching model's surprise will be identical to the other models if the final model it ends up switching to is the other model.

\pgfplotstableread[col sep=tab,]{data/winningsKL_T20_overall.csv}\datatable
\begin{table}
	\centering
	\pgfplotstabletypeset[
    	column type=c,
    	columns={Model,Full BToM, TWG, TW, TG, Switching},
     	every head row/.style={
     		before row={
     			\toprule
    		},
    		after row=\midrule,
    	},
    	every last row/.style={
    		after row=\bottomrule},
     	every column/.code={
            \ifnum\pgfplotstablecol>0
                \pgfkeysalso{clear infinite}
                \pgfkeysalso{column type/.add={|}{}}
                \pgfkeysalso{preproc/expr={100*##1}}
                \pgfkeysalso{postproc cell content/.append style={
                        /pgfplots/table/@cell content/.add={}{
                            \ifnum\pdfstrcmp{##1}{nan}=0
                                 --
                            \else
                                \%
                            \fi
                        },
                    }
                }
            \fi 
        },
        columns/Model/.style ={string type, column name=},
	col sep=&,row sep=\\,]{\datatable}
	\caption{Average frequency of the model on the left achieving a lower $S_1$ surprise measure than the model on the top across all trajectories.}
	\label{tab:winningsOverall}
\end{table}

\pgfplotstableread[col sep=tab,]{data/winningsRel_T1.5_overall.csv}\datatable
\begin{table}
	\centering
	\pgfplotstabletypeset[
    	column type=c,
    	columns={Model,Full BToM, TWG, TW, TG, Switching},
    	every head row/.style={
     		before row={
     			\toprule
    		},
    		after row=\midrule,
    	},
    	every last row/.style={
    		after row=\bottomrule},
    	every column/.code={
            \ifnum\pgfplotstablecol>0
                \pgfkeysalso{clear infinite}
                \pgfkeysalso{column type/.add={|}{}}
                \pgfkeysalso{preproc/expr={100*##1}}
                \pgfkeysalso{postproc cell content/.append style={
                        /pgfplots/table/@cell content/.add={}{
                            \ifnum\pdfstrcmp{##1}{nan}=0
                                --
                            \else
                                \%
                            \fi
                        },
                    }
                }
            \fi 
        },
        columns/Model/.style ={string type, column name=},
	col sep=&,row sep=\\,]{\datatable}
	\caption{Average frequency of the model on the left achieving a lower $S_2$ surprise measure than the model on the top across all trajectories.}
	\label{tab:winningsOverallRel}
\end{table}

Table \ref{tab:accuracyAggregated} suggested that the \emph{Full BToM} model outperforms the \emph{TW} model in the \emph{Destination Uncertainty} condition.
However, looking at the relative wins for the \emph{Destination Uncertainty} condition in more detail (Table \ref{tab:winningsTW}), we actually find a slightly different result compared to column \emph{DU} from Table \ref{tab:accuracyAggregated}:

The \emph{TW} model actually has a better accuracy in slightly over 58\% of the trajectories in the \emph{Destination Uncertainty} condition. The better surprise score shown in Table \ref{tab:accuracyAggregated} can be understood by taking into account how much worse a model performs in case it looses. Whenever the \emph{TW} model looses against the \emph{Full BToM} model, it performs a lot worse than when the \emph{Full BToM} model looses against it.

\pgfplotstableread[col sep=tab,]{data/winningsKL_T20_2.csv}\datatable
\begin{table}
	\centering
	\pgfplotstabletypeset[
    	column type=c,
    	columns={Model,Full BToM, TWG, TW, TG, Switching},
     	every head row/.style={
     		before row={
     			\toprule
    		},
    		after row=\midrule,
    	},
    	every last row/.style={
    		after row=\bottomrule},
     	every column/.code={
            \ifnum\pgfplotstablecol>0
                \pgfkeysalso{clear infinite}
                \pgfkeysalso{column type/.add={|}{}}
                \pgfkeysalso{preproc/expr={100*##1}}
                \pgfkeysalso{postproc cell content/.append style={
                        /pgfplots/table/@cell content/.add={}{
                            \ifnum\pdfstrcmp{##1}{nan}=0
                                 --
                            \else
                                \%
                            \fi
                        },
                    }
                }
            \fi 
        },
        columns/Model/.style ={string type, column name=},
	col sep=&,row sep=\\,]{\datatable}
	\caption{Average frequency of the model on the left achieving a lower $S_1$ surprise measure than the model on the top across all trajectories of the \emph{Destination Uncertainty} condition.}
	\label{tab:winningsTW}
\end{table}

\subsection{Computational Efficiency}

The second aspect for satisficing mentalizing is concerned with the computational costs of the different models. 
All models were implemented in the same way in Python using the same underlying data structure for the maze and the behavioral data.
The only differences between the different models arise from the different mental states that need to be considered for the different models resulting in slightly different likelihoods, most notably in the use of the actual distance versus the assumed distance according to the freespace assumption when considering the world belief.
The switching models makes use of the specialized models unmodified. 

The models not using the freespace assumption can improve their computational costs by  caching the exact distances after computing them once for any given start and end position. 
As the assumed distance under the freespace assumption does not only depend on the start and end position, but also on the already seen grids in the world of that trajectory, caching is not as straight forward.
In order to make the comparison more fair, we reset the cache after every trajectory, meaning that the models can at most benefit from it when a trajectory revisits an earlier position, for the results reported below. 

We measured the wall-time of all models for all trajectories and computed the averages across the different conditions as well as overall, similar to our accuracy scores above.
Table \ref{tab:timingsAbs} presents the results of this using $S_1$ on a 3.5Ghz Xeon machine using Python 3.6.8.

\accuracyTable{data/timingsAggregatedKL_T20.csv}{tab:timingsAbs}{Absolute average timings and their standard deviations (in ms) for the different models and conditions when using $S_1$.}

The models are ordered as expected from the least complex \emph{TWG} model taking the least amount of time to the \emph{Full BToM} model being orders of magnitude slower. 
Due to the freespace assumption the \emph{TG} model actually considers less mental states than the \emph{TW} model, which results in slightly better wall times.

In order to better highlight the differences we normalized these times by dividing them by the times of the quickest model (usually the \emph{TWG} model, except for the \emph{No Uncertainty} condition using $S_1$). 
These relative times are reported in Table \ref{tab:relTimings}. 
We also included the relative timings using the $S_2$ measurement here to show the relative differences. 
All models except the \emph{Full BToM} model are influenced in a similar way by the more complex computations involved in $S_2$. 
The \emph{Full BToM} model, however, suffers noticeably more.

The relative timings clearly show the different computational complexities for the different models with the specialized models being more efficient than the more general \emph{Full BToM} and the \emph{Switching} model.
The \emph{Switching} model is however an order of magnitude more efficient than the \emph{Full BToM} model, despite the very naive re-evaluation strategy currently employed. 

 
\pgfplotstableread[col sep=tab,]{data/timingsAggregatedKL_T20.csv}\datatable
\pgfplotstableread[col sep=tab,]{data/timingsAggregatedRel_T1.5.csv}\datatableRel
\pgfplotstablecreatecol[
	create col/copy column from table={\datatableRel}{overallRel}
	]{overallRel2}{\datatable}
\pgfplotstablecreatecol[
	create col/copy column from table={\datatableRel}{twgRel}
	]{twgRel2}{\datatable}
\pgfplotstablecreatecol[
	create col/copy column from table={\datatableRel}{twRel}
	]{twRel2}{\datatable}
\pgfplotstablecreatecol[
	create col/copy column from table={\datatableRel}{tgRel}
	]{tgRel2}{\datatable}

\begin{table}
	\centering
	\pgfplotstabletypeset[
	    zerofill,
    	column type=c,
    	columns={Model,overallRel,twgRel,twRel,tgRel,overallRel2,twgRel2,twRel2,tgRel2},
     	every head row/.style={
     		before row={
     			\toprule
    		},
    		after row=\midrule,
    	},
    	every last row/.style={
    		after row=\bottomrule},
     	every column/.code={
            \ifnum\pgfplotstablecol>0
                \pgfkeysalso{column type/.add={|}{}}
            \fi
        },
        columns/Model/.style ={string type},
        columns/overallRel/.style ={column name=Overall},
        columns/twgRel/.style ={column name=NU},
        columns/twRel/.style ={column name=DU},
        columns/tgRel/.style ={column name=PU},
        columns/overallRel2/.style ={column name=Overall $S_2$, column type/.add={|}{}},
        columns/twgRel2/.style ={column name=NU $S_2$},
        columns/twRel2/.style ={column name=DU $S_2$},
        columns/tgRel2/.style ={column name=PU $S_2$},
	col sep=&,row sep=\\,]{\datatable}
	\caption{Relative times of the different models using $S_1$ and $S_2$, normalized to the quickest model.}
	\label{tab:relTimings}
\end{table}

Since the time for the \emph{Switching} model heavily depends on the number of re-evaluations, due to the naivety of our initial approach, we also report the average number of re-evaluations when using $S_1$ for all variations in our data in Table \ref{tab:switches}.
Here we find that the \emph{No Uncertainty} condition unsurprisingly causes the least amount of re-evaluations as the \emph{Switching} model can usually stick to its initial \emph{TWG} model. 
In the other conditions, we find that the number of re-evaluation is dependent on the maze. In particular the length of the optimal trajectory (which is longest for Maze 1,3 and 4) influences the number of re-evaluation which is understandable, when considering that $S_1$ will almost always increase due to how it is computed. 
Maze 6 is comparatively speaking shorter in terms of optimal distance, however it appears to be more tricky for participants (cf. average distance in Table \ref{tab:optimality}), resulting in the \emph{Switching} model to have to re-evaluate more often as well.

\pgfplotstableread[col sep=semicolon,]{data/avg_switchesKL_T20.csv}\datatable
\begin{table}
	\centering
	\pgfplotstabletypeset[
	    precision=3,
    	column type=c,
    	columns={Maze,C1V1,C1V2,C2V1,C2V2,C3V1,C3V2},
     	every head row/.style={
     	    before row={
     	        \toprule
				& \multicolumn{2}{c|}{NU} & \multicolumn{2}{c|}{DU} & \multicolumn{2}{c}{PU}\\
			},
     		after row=\midrule,
     	},
    	every last row/.style={
    		after row=\bottomrule},
     	every column/.code={
            \ifnum\pgfplotstablecol>0
                \pgfkeysalso{column type/.add={|}{}}
            \fi
        },
    columns/Maze/.style={},
	columns/C1V1/.style={column name=V1, /pgf/number format/fixed, fixed zerofill},
	columns/C1V2/.style={column name=V2, /pgf/number format/fixed, zerofill},
	columns/C2V1/.style={column name=V1},
	columns/C2V2/.style={column name=V2},
	columns/C3V1/.style={column name=V1},
	columns/C3V2/.style={column name=V2},
	col sep=&,row sep=\\,]{\datatable}
	\caption{Relative number of re-evaluations performed by the switching model across the different mazes and conditions using $S_1$ with an initial threshold of 20.}
	\label{tab:switches}
\end{table}

\subsection{Exemplary Model Traces}

To demonstrate how the models (the proposed \emph{Switching} model in particular) react to the different observations during a particular trajectory.

\begin{figure}
	\centering
	\begin{subfigure}[c]{\textwidth}
	    \centering
		\includegraphics[width=0.48\textwidth]{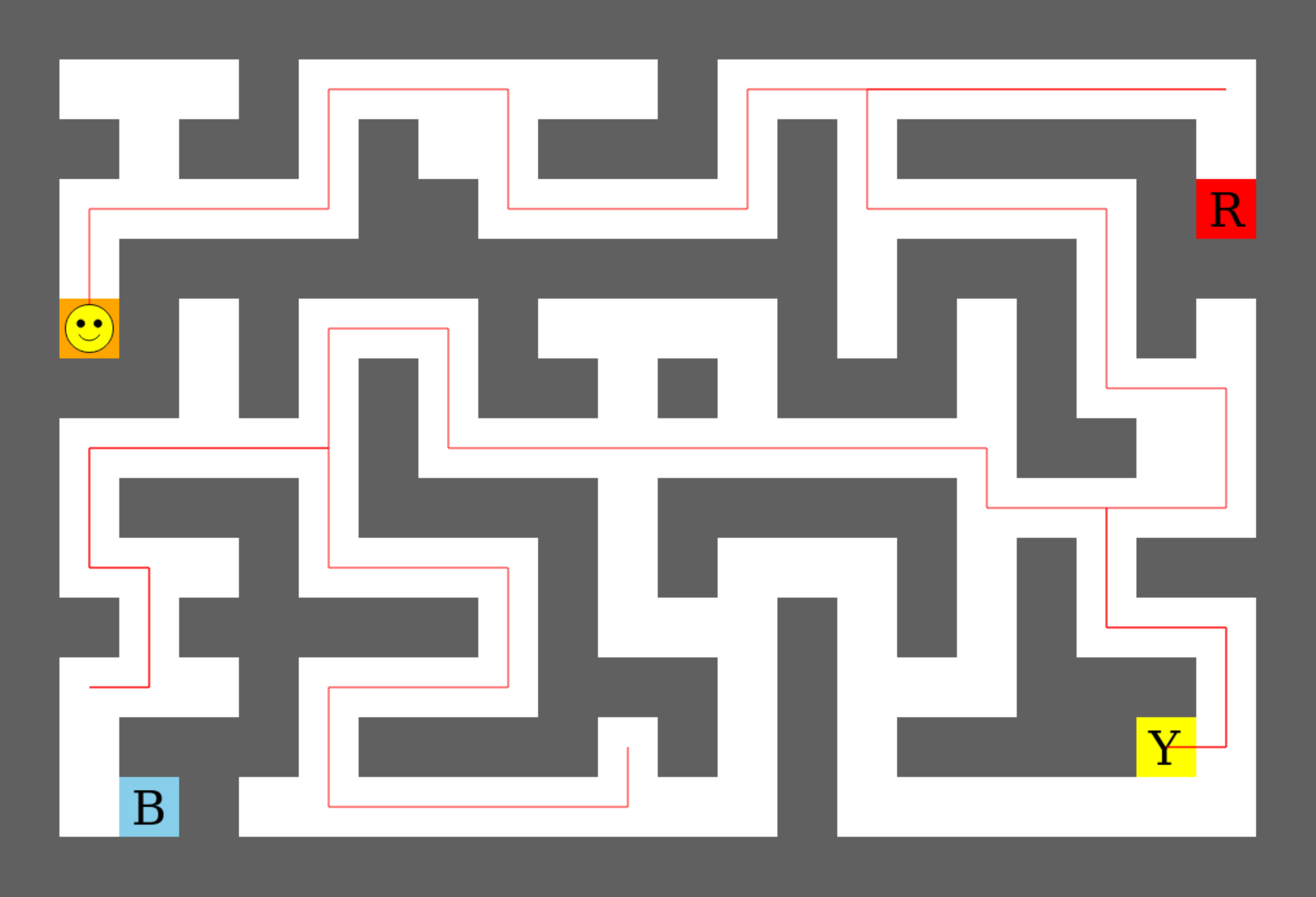} 
		\hfill
		\includegraphics[width=0.48\textwidth]{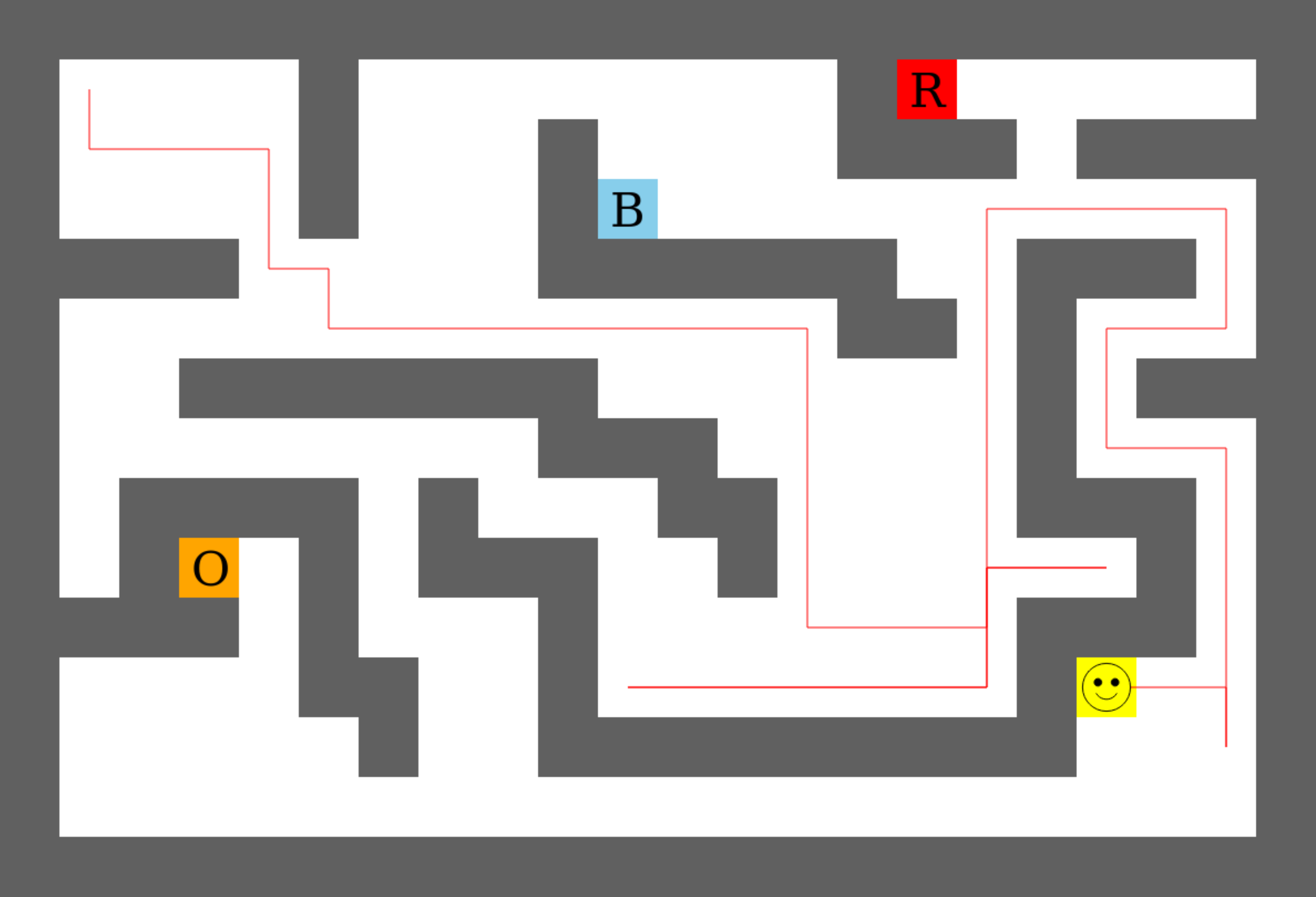} 
		 \subcaption{Example trajectory in the DU condition in Maze 3 (left) and  the PU condition in Maze 6 (right).}
		 \label{fig:examples}
	\end{subfigure}
	\begin{subfigure}[c]{\textwidth}
		\pgfplotstableread[col sep=semicolon,]{data/scoreListP48_321_KL_20.csv}\dataOnecsv
		\pgfplotstableread[col sep=semicolon,]{data/scoreListP29_631_KL_20.csv}\datacsv
		\begin{tikzpicture}
		\begin{groupplot}[
		ymin=0,
		xmin=0,
		width=0.5\textwidth,
		group style = {group size = 2 by 1, 
			horizontal sep = 45pt, 
		},
		ylabel=$S_1$ Score,
		xlabel=Steps,
		every axis plot/.append style={very thick}
		]
		\nextgroupplot[
		legend style = { column sep = 10pt, legend columns = -1, legend to name = scoreLegend,},
		cycle list name=linestyles*,
		]
		\addplot table [x=Step, y=Full BToM]{\dataOnecsv}; 
		\addlegendentry{Full BToM};
		\addplot table [x=Step, y=TWG]{\dataOnecsv};
		\addlegendentry{TWG};
		\addplot table [x=Step, y=TW]{\dataOnecsv};
		\addlegendentry{TW};
		\addplot table [x=Step, y=TG]{\dataOnecsv};
		\addlegendentry{TG};
		\addplot+ [red] table [x=Step, y=Switching]{\dataOnecsv}; \addlegendentry{Switching};
		\addplot[ycomb,gray] table[x=Switches,y expr=110] {\dataOnecsv};
		\nextgroupplot[
		cycle list name=linestyles*,]
		\addplot table [x=Step, y=Full BToM]{\datacsv};
		\addplot table [x=Step, y=TWG]{\datacsv};
		\addplot table [x=Step, y=TW]{\datacsv};
		\addplot table [x=Step, y=TG]{\datacsv};
		\addplot+ [red] table [x=Step, y=Switching]{\datacsv};
		\addplot[ycomb,gray] table[x=Switches,y expr=110] {\datacsv};
		\end{groupplot}
		\node at ($(group c2r1) + (-4.5cm,-3.9cm)$) {\ref{scoreLegend}}; 
		\end{tikzpicture}
		\subcaption{Negative log likelihood ($S_1$) over steps for the different models for the trajectories shown in \ref{fig:examples}. Vertical lines indicate re-evaluations by the \emph{Switching} model.}
		\label{fig:scores}
	\end{subfigure}
	\begin{subfigure}[c]{\textwidth}
		\pgfplotstableread[col sep=semicolon,]{data/entropyP48_321_KL_20.csv}\dataOnecsv
		\pgfplotstableread[col sep=semicolon,]{data/entropyP29_631_KL_20.csv}\datacsv
		\begin{tikzpicture}
		\begin{groupplot}[
		ymin=0,
		xmin=0,
		width=0.5\textwidth,
		height=0.27\textheight,
		group style = {group size = 2 by 1, 
			horizontal sep = 45pt, 
		},
		ylabel=Average Entropy,
		xlabel=Steps,
		every axis plot/.append style={very thick}
		]
		\nextgroupplot[
		cycle list name=linestyles*,
		]
		\addplot table [x=Step, y=Full BToM]{\dataOnecsv}; 
		\addplot table [x=Step, y=TWG]{\dataOnecsv};
		\addplot table [x=Step, y=TW]{\dataOnecsv};
		\addplot table [x=Step, y=TG]{\dataOnecsv};
		\addplot+ [red] table [x=Step, y=Switching]{\dataOnecsv}; 
		\nextgroupplot[
		cycle list name=linestyles*,]
		\addplot table [x=Step, y=Full BToM]{\datacsv};
		\addplot table [x=Step, y=TWG]{\datacsv};
		\addplot table [x=Step, y=TW]{\datacsv};
		\addplot table [x=Step, y=TG]{\datacsv};
		\addplot+ [red] table [x=Step, y=Switching]{\datacsv};
		\end{groupplot}
		\end{tikzpicture}
		\caption{Averaged entropy across the three considered mental states for the different models.}
		\label{fig:entropy}
	\end{subfigure}
	\caption{Example trajectories \ref{fig:examples} and the corresponding output of the different models: \ref{fig:scores} action predictions, \ref{fig:entropy} entropy of inferred mental states. Left: \emph{Destination Uncertainty} condition; right: \emph{Path Uncertainty} condition}
	\label{fig:scoreListExamples}
\end{figure}
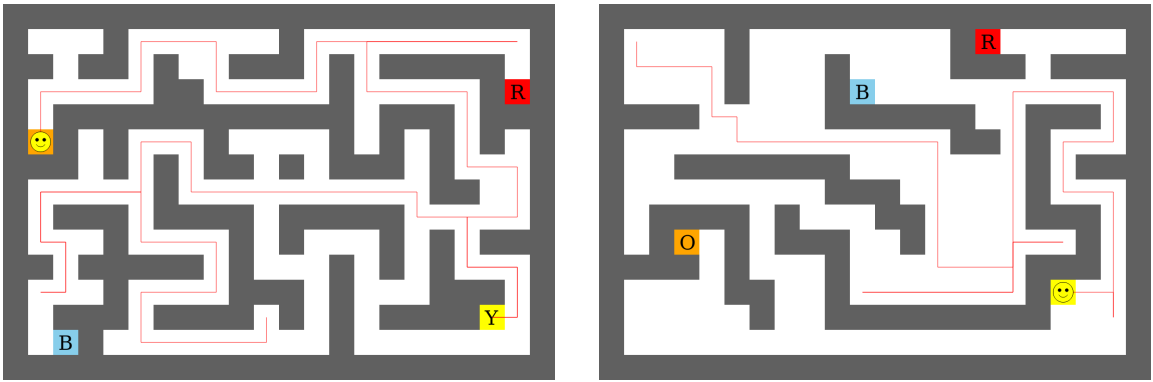
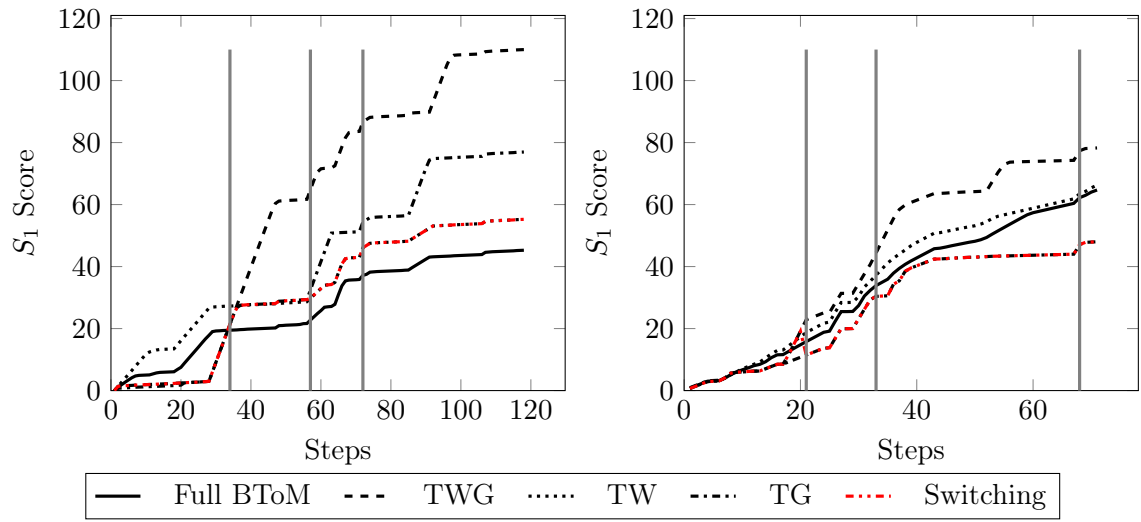
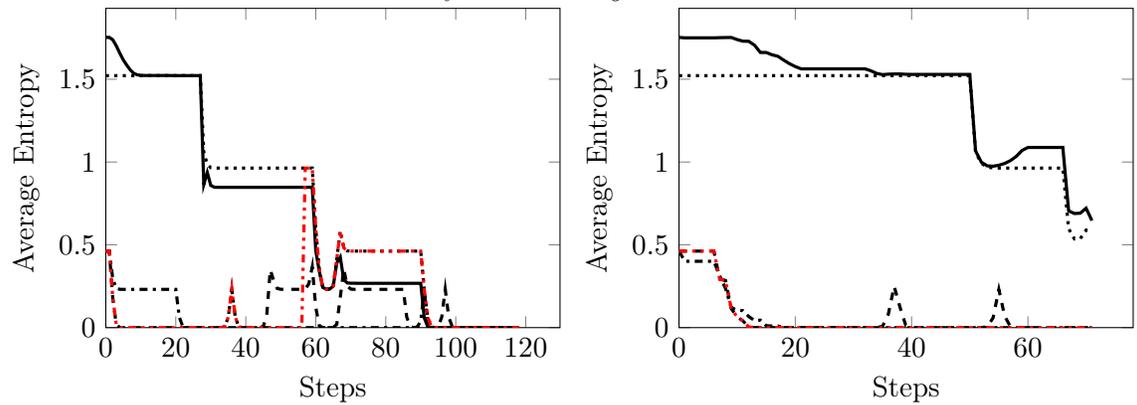

Figure \ref{fig:scoreListExamples} shows two example trajectories, one from the \emph{Destination Uncertainty} condition in Maze 3 (left side of \ref{fig:examples}) and one from the \emph{Path Uncertainty} condition in Maze 6 (right side).
In Figure \ref{fig:scores} we present the models scores across the different steps within the trajectories. These scores correspond to the negative log-likelihood ($S_1$). 
For the left image, we see that the \emph{Switching} model quickly switches to the \emph{TW} model at step 34 and sticks to it as it explains the behavior best. The \emph{Full BToM} model however, can explain the behavior even better in terms of $S_1$ primarily due to a better start in the first 10 steps.
For the right image (the \emph{Path Uncertainty} condition), the \emph{Switching} model correctly switches to the \emph{TG} model at step 21. 

The plots in Figure \ref{fig:entropy} show how certain the different models are regarding the inferred mental states. 
We computed the entropy for all three mental states (desire, goal belief and world belief) and averaged them after each step. Obviously the \emph{Full BToM} model has a strong disadvantage here, as it considers the most mental states.
The initial difference between the \emph{TW} and \emph{TG} model comes from the fact that our simplification using the freespace assumption results in the \emph{TG} model to only consider desires, whereas the \emph{TW} model additionally considers the goal beliefs with 24 different outcomes.
The plots show that the \emph{Full BToM} and \emph{TW} models only really improve their entropy upon seeing any of the exits since they cannot rule out any world before that. 
For the \emph{Destination Uncertainty} condition example on the left, we find that the \emph{Full BToM} and the \emph{TW} models can actually reduce their entropy to zero after seeing the 3rd exit at step 91 and then moving away from it, as they will be certain about the world they are in at that point. 
When they do not see all exits, as in the \emph{Path Uncertainty} example on the right, both models will not be able to eliminate their uncertainties. They will not even be able to make desire predictions until they observe the agent seeing the final exit and walking towards it (around step 68).

\section{Discussion \label{sec:discussion}}

The contributions of this paper are twofold: 
First of all, we want to draw attention to the benefits and deficits of BToM models of varying complexity. 
Our results clearly show that specialized and simpler models are usually better at predicting an agent's behavior compared to more complex models, as long as the models assumptions are not violated.
The more flexible and powerful \emph{Full BToM} model has too much flexibility. 
For large parts of a behavior, it will not be able to make confident predictions as it will not be able to rule out certain combinations of desires and mental states, making almost any action equally likely (cf. Figure \ref{fig:entropy}). 
On the other hand, the simpler models are able to make predictions more easily, as they do not consider as many possibilities due to their discrete assumptions regarding certain mental states.
When these assumptions are violated, we unsurprisingly find that these models perform a lot more poorly. 
These findings highlight, on the one hand side, that specialized models for very specific scenarios will not suffice for artificial systems trying to interact with humans. 
On the other hand side, however, a general, all-encompassing model will most likely also not be able to yield satisfying results. 
While a general model will rarely make wrong predictions or inferences, it will likewise not make any strong predictions either, reducing its usefulness for many scenarios. We have also found that humans tend to be rather quick at making discrete inferences, when not primed to consider multiple possible alternatives \cite{poeppelCogSci2019}.
Human evolution may also favour stronger predictions and inferences, which may further explain the emergence of heuristics in human reasoning \cite{lieder_griffiths_2019,haselton2015evolution}.

On top of these accuracy considerations, our results also show the well known computational complexity inherent in Bayesian models.
The \emph{Full BToM} model can be considered as a combination of the \emph{TW} and \emph{TG} model, but its complexity rises exponentially, even in our simple discrete 2D domain.
This results in the \emph{Full BToM} model are orders of magnitude slower than the specialized models, especially, when compared to the simplest model, which by itself already predicts an agent's next actions fairly well.
While one can combat this exponential explosion using approximate inference methods, this always introduces a different trade-off between accuracy and speed. 
Furthermore, while the benefits of approximate methods will not apply linearly to all models, the simpler models will also benefit from them.
As such we see approximate methods as an alternative approach towards satisficing mentalizing compared to our proposed \emph{Switching} model, which is the focus of the second contribution of this paper:

While very simple, the proposed switching approach combines the benefits of the specialized models and the flexibility of the \emph{Full BToM} model. 
Similar to the egocentric bias in humans \cite{keysar2007communication}, the \emph{Switching} approach will initially project its own knowledge onto the agent, thus using the simplest model of only inferring the other's desire. 
Only when it encounters counter-evidence against its current model choice, will it reconsider and switch to a more appropriate model. 
This approach was inspired by the processing steps found in humans upon encountering surprising events \cite{meyer1997toward}.
Even with our very naive re-evaluation strategy, we already outperform all other models in terms of predictive accuracy, since for all conditions, it can make use of the most appropriate model, even if the human behavior may deviate from what would be considered the norm for that condition.
The findings in Table \ref{tab:accuracy21} where the \emph{TG} model outperforms the \emph{Switching} approach in two conditions indicates that our meta parameters (primarily $\gamma$ in this case) can still be optimized further, when one strives for the best accuracy, however lower $\gamma$ would lead to more re-evaluations and thus worse efficiency.

At the same time it remains at least an order of magnitude quicker than the \emph{Full BToM} model since it can keep the number of re-evaluations down while never considering a model as complex as the \emph{Full BToM} model.
Another benefit of the \emph{Switching} approach with regard to satisficing mentalizing is its potential adaptivity. 
Satisficing is all about being \emph{good enough}, however this is not a static quantity. What is good enough will always depend on the individual situation and will be influenced by a range of different factors, such as available resources or one's current mental state. 
Any number of such factors can easily be incorporated into the \emph{Switching} approach by adapting the threshold for the re-evaluation appropriately. 
Possible factors an artificial system could take into account involve the current trade-off between accuracy and speed, learned success of available models in past encounters, and/or the cost of actually performing the re-evaluation - something we are currently looking into in ongoing work.
Furthermore, one can not only adapt the point in time when a re-evaluation should happen, but also how it should happen. 
Currently, we naively evaluate all possible models, an approach not really suitable once the number of considered models increases. 
However, one could also imagine a hierarchical re-evaluation mechanism which has been learned and adapted over time.
Finally, the \emph{Switching} approach can easily be extended online by adding new specialized models to the pool of considered models. This is similar to the learning of heuristics in humans \cite{lieder_griffiths_2019}.

A third aspect we want to highlight in this paper is the need to evaluate mentalizing models on actual human behavior data. 
Human behavior can deviate quite strongly from the behavior assumed to be the norm given a specific situation. 
In our data, we not only found behavior in the \emph{Path Uncertainty} condition being very similar to the \emph{No Uncertainty} condition on certain mazes, but also found behavior in the \emph{Destination Uncertainty} condition that can not be explained by the assumed mental states of that condition. 

\section{Conclusion}

In this paper we have presented a \emph{Switching} strategy as an alternative to a Full Bayesian Theory of Mind model to perform mentalizing in scenarios where the observed agents may have a range of different mental states. 
The Switching strategy proposes the use of specialized models, designed to handle different mental potential mental states in isolation, but only consider one of these models at any given time, switching between them as appropriate, similar to the use of different learned heuristics in humans \cite{lieder_griffiths_2019}.
We use a measurement of \emph{surprise} to make this decision, modelled after the adaptive processes in humans upon encountering unexpected or surprising events \cite{meyer1997toward}.

We evaluated our proposed approach not as commonly done via correlations with human judgements on synthetic data but rather their predictive accuracy on actual human behavior data, which we collected in a study where we manipulated the available information participants had to work with.
While comparing ToM models with human ratings is important, we believe that looking at actual human data and considering predictive performance should not be neglected. 
After all one can argue that the actual inferred mental state may not matter as long as the resulting behavior of the system remains optimal \cite{pynadath2007minimal}.
Our results show not only that specialized models can outperform more complex ones in situations where their assumptions are valid, but also that our proposed \emph{Switching} approach can outperform a general BToM model in terms of both predictive accuracy as well as computational complexity.
The benefits of stronger predictive power as well as lower computational costs of specialized models can be employed across different scenarios. 
A more complex model is only used when required, which is especially useful at the beginning of the behavior, where more complex models cannot rule out many of their possibilities. 

Furthermore, while we did not evaluate different switching thresholds and re-evaluation strategies in this paper, we believe that the general approach is a good step towards satisficing mentalizing for artificial systems. 
We do believe that the current reconsideration strategy needs improvement, as evaluating all specialized models will not scale to more complex scenarios with more of these models. 
Apart from learning a separate model which will predict the most useful model, once could also consider only evaluating the different models at a subset of the seen behavior, determined by which actions have been surprising before.
Another alternative may be a predetermined hierarchy of models which is evaluated in the hierarchical order.
Regardless of the final strategy, the easy inclusion of adaptive thresholds and/or dynamic re-evaluation strategies allows this approach to yield results that are \emph{good enough} for a given situation, while conforming to computational constraints.

While sampling may also serve to reduce the computational demands of BToM models and may even be useful to mirror human fallacies \cite{vul2014one}, we see our approach as an alternative. 
In fact, depending on the employed switching strategy and considered specialized models, one can formulate the \emph{Switching} approach as a sampling strategy when one re-samples discrete mental states within the \emph{Full BToM} model. 
However, in this case, one needs to find a way to update the prior probabilities for the different mental states according to the observations efficiently. 
We are currently exploring some ways of how this could be achieved in ongoing work (e.g. \cite{poeppelCogSci2019}).
Another important avenue of improving this approach is by allowing for dynamic learning of specialized models. Currently, all considered models need to be specified a priori as in the BToM framework. 
Ideally, the system would realize when none of its models can explain the observed behavior to a satisfying degree and start adapting existing models as well as creating new ones. 

\vskip 0.2in
\bibliography{jpoeppelSatisficing}
\bibliographystyle{theapa}

\end{document}